%% file: main.tex
\documentclass[10pt,twocolumn,letterpaper]{article}

\usepackage[pagenumbers]{cvpr} % To force page numbers, e.g. for an arXiv version

\definecolor{cvprblue}{rgb}{0.21,0.49,0.74}
\usepackage[pagebackref,breaklinks,colorlinks,allcolors=cvprblue]{hyperref}
\usepackage{multirow}
\usepackage{subcaption}
\usepackage{enumitem}

\usepackage{multicol}

\usepackage{listings}
\usepackage{xcolor}
\usepackage{multirow}
\usepackage{subcaption}
\usepackage{xcolor}
\usepackage{hyperref}

\newif\ifhighlight
\highlightfalse

\ifhighlight
    \newcommand{\reviseH}[1]{\textcolor{orange}{#1}}
    \newcommand{\rewriteH}[1]{\textcolor{blue}{#1}}
\else
    \newcommand{\reviseH}[1]{#1}
    \newcommand{\rewriteH}[1]{#1}
\fi

\title{AgentHOI: Multi-\underline{Agent} Reasoning for \underline{H}uman-\underline{O}bject-\underline{I}nteraction Video Generation via Implicit Representation Alignment}

\author{
Ziyao Huang$^{1,2}$,~Shunkai Li$^2$,~Juan Cao$^1$,~Chengyu Li$^1$,~Youliang Zhang$^2$,\\
~Zixiang Zhou$^2$,~Cong Wang$^2$,~Yuan Zhou$^2$\protect\footnotemark[2],~Qinglin Lu$^2$,~Fan Tang$^1$\protect\footnotemark[1]\\
$^1$ University of Chinese Academy of Sciences
~ $^2$ Tencent HunYuan
}

\begin{document}
% \maketitle
\input{figures/teaser}

\begingroup
\renewcommand{\thefootnote}{\fnsymbol{footnote}}
\footnotetext[2]{Project leader: Yuan Zhou.}
\footnotetext[1]{Corresponding author: Fan Tang.}
\endgroup

\input{sec/0_abstract}
\input{sec/1_intro}
\input{sec/2_relatedwork}
\input{sec/3_method}

\input{sec/4_expr}
\input{sec/5_conclusion}
{
    \small
    \bibliographystyle{ieeenat_fullname}
    \bibliography{main}
}

% WARNING: do not forget to delete the supplementary pages from your submission 
\clearpage
\input{sec/X_suppl}

\end{document}

%% file: figures/teaser.tex
\twocolumn[{%
\renewcommand\twocolumn[1][]{#1}%
\maketitle
\begin{center}
    %\centering
    \captionsetup{type=figure}
  \includegraphics[width=\textwidth]{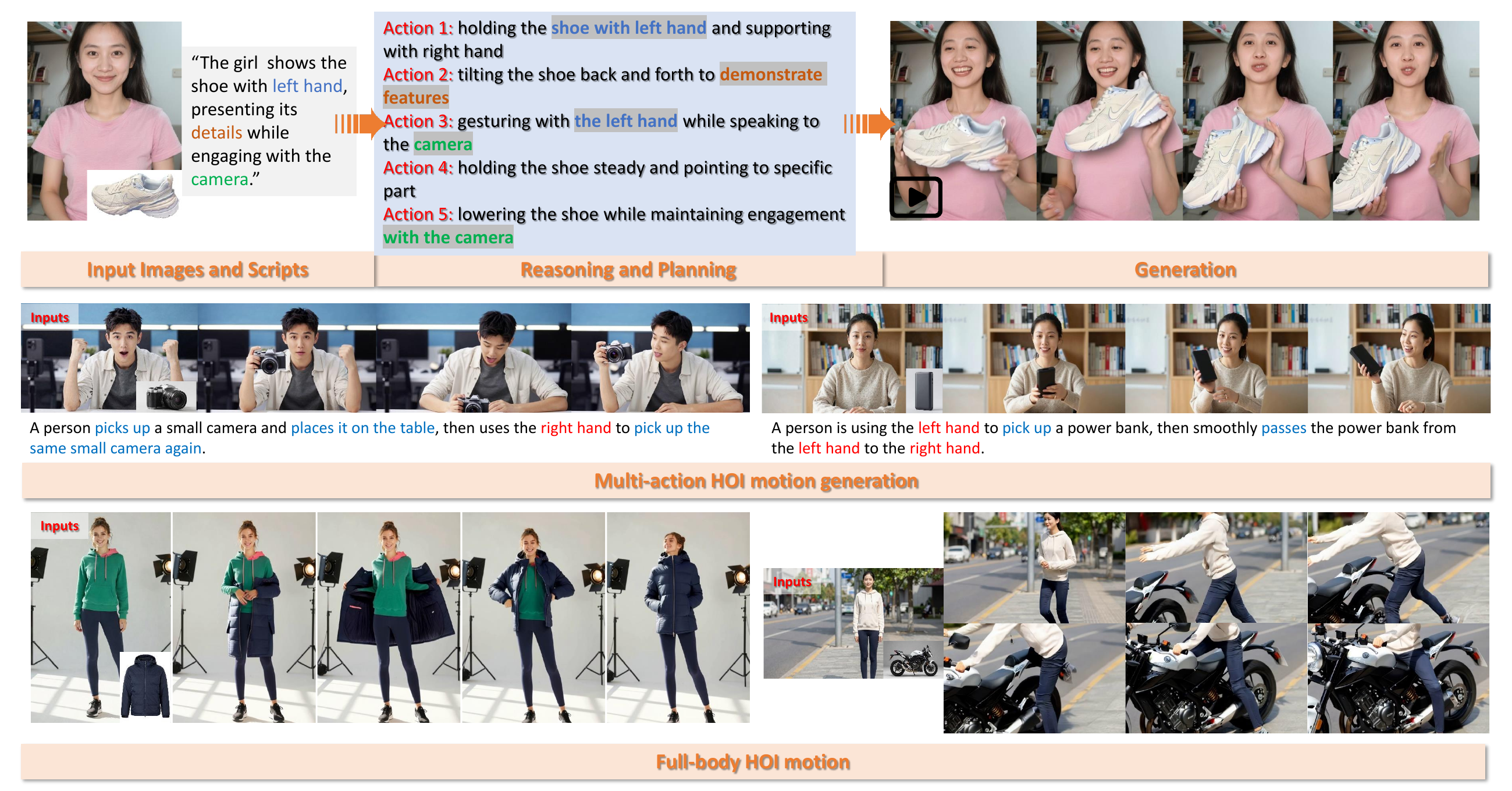}
  \caption{Following a ``thinking-before-generation'' schema, the proposed AgentHOI first generates detailed textual motion controls and then generates corresponding HOI videos.
  The Action 1--Action 5 labels in the teaser are simplified for visualization; the actual motion plan uses second-level temporal intervals.
  }
  \label{fig:teaser}
\end{center}%
}]

%% file: sec/0_abstract.tex
\begin{abstract}
Recent advances in video diffusion models have spurred interest in human–object interaction (HOI) video generation, which demands fine-grained control over interaction logic beyond single-subject animation. 
However, existing HOI methods rely heavily on explicit motion control, limiting scalability and generalization across diverse objects and interactions.
In this study, we propose AgentHOI, a text-driven HOI video generation following a thinking-before-generation framework that bridges the gap between high-level textual intent and physical execution through multi-agent reasoning over perception, interaction, and motion planning. 
Building upon the generated interaction plans, we further strengthen text-driven motion understanding. We introduce an implicit text–motion alignment strategy that distills text-to-motion priors into the video diffusion model, enabling robust HOI synthesis without explicit motion inputs at inference.
Experiments show that AgentHOI significantly improves interaction naturalness, object appearance preservation, and adherence to complex textual instructions across challenging object-centric scenarios such as wearing and riding.
The code is available at \url{https://github.com/bone-11/agenthoi}.
\end{abstract}

%% file: sec/1_intro.tex
\section{Introduction}
\label{sec:intro}

With the advancement of video diffusion models~\cite{wang2025wan, kong2024hunyuanvideo}, the quality of digital human animation~\cite{hu2024animateanyone, zhang2024mimicmotion} has been significantly improved. 
As the field matures, the research frontier is rapidly shifting from animating isolated human subjects to modeling complex Human-Object Interactions (HOI)~\cite{xu2024anchorcrafter, xue2024hoi, pang2025manivideo, huang2025homa, wang2025dreamactor-h1}, demonstrating strong potential for applications such as e-commerce, digital advertising, and film production.
However, unlike single-subject animation, HOI video generation necessitates a profound understanding of interactive motion: the subtle interplay between human intent, hand-object contact, and the physical response of the object.

Current methods typically rely on explicit motion control signals to generate human-and-object actions, such as motion capture~\cite{xu2024anchorcrafter, fan2025rehold, huang2025homa} or computationally expensive 3D modeling~\cite{pang2025manivideo, wang2025dreamactor-h1}.
While showing some promising results, these approaches suffer from severe limitations in real-world scalability: the acquisition of motion sequences for diverse objects is prohibitively expensive, and the models exhibit limited generalization to novel object categories. 
Furthermore, when encountering objects with complex geometries, deformable structures, or unique material properties (e.g., textiles or fluid containers), predefined explicit control signals often fail to capture the nuanced physical dynamics inherent in authentic interactions, frequently resulting in artifacts such as geometric penetration or unrealistic motion trajectories.

Inspired by the widespread success of text-to-video (T2V) generation in open-domain scenarios, text-driven control has emerged as a flexible and highly accessible interface for HOI video generation.
However, user-provided text primarily conveys high-level intent rather than precise motion execution details, remaining inherently underspecified when applied to the granular physics of HOI.
A simple prompt leaves critical execution detail entirely to the model's stochastic ``guesswork''.
This profound semantic gap between high-level intent and low-level pixel dynamics often causes diffusion models to collapse toward the statistical mean, manifesting as physically jarring artifacts like hand-clipping, floating objects, or stiff motion.
In this paper, we propose AgentHOI, a novel framework that bridges this gap through a ``thinking-before-generation'' schema. 
By introducing multi-agent reasoning, AgentHOI deconstructs complex HOI tasks into physically grounded execution blueprints before the first pixel is even synthesized.  

Specifically, our method is built upon three complementary components.
First, we formulate the HOI synthesis as a multi-agent orchestration problem, where specialized agents for \textbf{visual perception}, \textbf{relation analysis}, and \textbf{motion planning} collaborate to translate abstract prompts into temporally explicit action timelines.
This formulation establishes an explicit intermediate reasoning layer that mediates between high-level textual intent and executable motion.
Second, to improve text-driven controllability, we propose an implicit text-to-motion (T2M) representation alignment strategy, improving controllability while eliminating the need for explicit motion inputs during inference.
By utilizing a token relation distillation (TRD) item, we align the internal spatial and temporal token relationships of the T2V backbone with a T2M prior, enabling the model to internalize interaction dynamics.
Third, existing datasets are insufficient to support this task due to limited availability, high acquisition cost, and sparse coverage. To address these limitations, we develop a mixed-source data pipeline that combines real-world and synthetic data, along with a tailored training strategy that accounts for the distinct characteristics of synthetic data.

Experimental results on cross and self-motion driven test sets demonstrate that AgentHOI consistently outperforms SOTAs in text-driven HOI generation, achieving superior object appearance preservation, motion smoothness, and adherence to complex multi-step action instructions. In summary, our contributions are:
\begin{itemize}[topsep=0em,leftmargin=1.5em,itemindent=0em,labelsep=1em]
    \item We propose an agent-enhanced reasoning framework for HOI video generation that enables fine-grained and semantically consistent control of complex human–object interactions directly from text, overcoming the limited controllability of prior methods that rely on explicit motion representations.

    \item We develop a unified training framework that combines implicit text–motion–video feature alignment with a hybrid real-and-synthetic HOI data pipeline, leveraging text-to-motion priors to bridge high-level interaction semantics and video generation, and facilitating robust learning of text-controllable interaction dynamics across diverse objects and actions.

    \item Extensive experiments demonstrate that our method significantly improves text controllability, interaction naturalness, and object appearance preservation, and supports a wide range of challenging HOI applications that are difficult to model with explicit motion control, such as wearing, riding, and other object-centric interactions.

\end{itemize}

%% file: sec/2_relatedwork.tex
\section{Related Work}
\label{sec:related_work}

\subsection{Motion-controllable Human Video Generation}
With the rapid advancement of video generation models~\cite{blattmann2023svd, kong2024hunyuanvideo, wang2025wan, zhou2024allegro}, controllable human video synthesis has attracted increasing attention. A dominant line of research relies on explicit motion representations—such as pose sequences or audio-derived motion cues—to guide generation while preserving identity and appearance consistency.
For pose-driven control, prior works range from UNet-based architectures~\cite{hu2024animateanyone, zhu2024champ, wang2024humanvid, zhang2024mimicmotion, tu2024stableanimator, wang2024unianimate, huang2024makeyouranchor, AnimateX2025} to more recent DiT-based formulations~\cite{shao2024human4dit, wang2025unianimatedit, cheng2025wananimate}, achieving improved motion fidelity and temporal coherence. Similarly, audio-driven approaches extend from head animation~\cite{zhang2023sadtalker, ji2024sonic, tian2024emo, chen2025echomimic} to semi-body~\cite{corona2024vlogger, meng2024echomimicv2, lin2024cyberhost} and full-body animation~\cite{chen2025hunyuanvideoavatar, lin2025omnihuman}.
Despite their progress, pose-driven methods rely heavily on explicit motion retargeting, which requires complex and fragile preprocessing to align motion signals across subjects and appearances, limiting robustness and scalability. Moreover, motion modeling is largely human-centric, with little consideration of objects or environments, making these approaches ill-suited for human–object interaction (HOI) scenarios that demand consistent contact and joint motion reasoning.
In contrast, AgentHOI removes pose retargeting entirely, deriving HOI dynamics from text via multi-agent reasoning and implicit motion-structure alignment during training.

\subsection{Human-object Interaction Video Generation}
As controllable human video generation advances, enabling human–object interaction (HOI) has become an important yet challenging direction. Prior work outside video generation focuses on HOI motion synthesis using explicit geometric or physical representations~\cite{jiang2024trumans, jiang2024autonomous, xu2023interdiff, li2023object, peng2023hoi, ghosh2023imos}, while HOComp~\cite{hocomp} performs interaction-aware human–object composition only at the image level. Multi-subject reference-conditioned video generation methods such as HuMo~\cite{chen2025humo} and Video Alchemist~\cite{videoalchemist} preserve multiple reference subjects but treat them independently without modeling the joint motion and contact patterns required by HOI. Within video generation, several methods preserve interactions in real videos by replacing humans or objects, such as MIMO~\cite{men2025mimo} and AnimateAnyone2~\cite{hu2025animateanyone2} for human substitution, and HOI-Swap~\cite{xue2024hoi}, ReHoLD~\cite{fan2025rehold}, and iDiTHOI~\cite{shen2025idithoi} for object replacement. However, these approaches are limited to editing existing content rather than generating HOI videos from scratch.

To synthesize HOI from scratch, existing methods predominantly rely on explicit motion control. AnchorCrafter~\cite{xu2024anchorcrafter} and ManiVideo~\cite{pang2025manivideo} model interactions using strong motion-related conditions, including pose, depth, or detailed 3D representations. More recent efforts attempt to relax motion constraints but remain motion-centric: SViMo~\cite{dang2025svimo} enables HOI generation through joint 2D–3D motion co-generation to align image-space and spatial dynamics, yet is restricted to hand–object interactions; HOMA~\cite{huang2025homa} introduces weak motion signals but still depends on motion capture or manual editing; and DreamActor-H1~\cite{wang2025dreamactor-h1} controls interactions via predefined HOI templates, limiting generalization.

Distinct from all the works above, AgentHOI generates HOI videos from scratch under purely textual control. A multi-agent HOI reasoning module decomposes user intent into a structured action timeline, and implicit text-motion representation alignment injects motion-structure priors during training, so no pose, depth, 3D, or motion-capture input is required at inference.

\subsection{Agent-based Human Video Generation}
A growing body of work in video generation introduces agent-based frameworks to improve generation quality through planning, reasoning, and iterative refinement~\cite{xie2024dreamfactory, hu2024storyagent, yuan2024mora, zheng2024videogenofthought, long2025vista, huang2024genmac}.
In the context of human video generation, several representative methods leverage agents to enhance motion controllability and semantic coherence. OmniHuman1.5~\cite{jiang2025omnihuman15} enables digital human motion control through explicit reasoning processes, while ORCA~\cite{he2025activeavatar} treats an Image-to-Video model as a world simulator, allowing an agent to infer world states and plan coherent action sequences to fulfill high-level intentions. SocialAgent~\cite{zhang2025socialagent} further incorporates agents into action generation to model high-level semantic behaviors.

However, in the domain of human–object interaction (HOI) video generation, the role of agents remains largely underexplored. This gap is particularly pronounced given the complex attribute dependencies and diverse interaction affordances between humans and objects, which pose challenges that go beyond human-centric motion planning and demand more structured, interaction-aware reasoning.
AgentHOI fills this gap with an HOI-specific agent pipeline that couples visual perception, relation analysis, motion planning, and clipping refinement, yielding a temporally structured action timeline that captures joint human–object affordances and contact.

%% file: sec/3_method.tex
\section{Method}

We formulate HOI video generation as animating a reference human image $I_H$ with an inserted object image $I_O$, conditioned on text prompts $T$, to produce an output video $Y = \{y^{1:n}\}$ consisting of $n$ frames. 
Unlike prior works~\cite{xu2024anchorcrafter, huang2025homa, wang2025dreamactor-h1} that necessitate explicit motion priors during inference, AgentHOI operates in a motion-free regime, deriving interaction dynamics through a hierarchical thinking-before-generation pipeline. 
The overall framework is illustrated in Fig.~\ref{fig:pipeline}, consisting of an agent-based HOI reasoning module (Sec.~\ref{sec:method.agent}) and the motion-free HOI video generation module (Sec.~\ref{sec:method.vdm}). 
To support training, the mixed-source data curation and training strategy is proposed in Sec.~\ref{sec:method.data}.

\input{figures/pipeline}

\subsection{Agent-based HOI Reasoning}
\label{sec:method.agent}
The objective of this module is to transform textual user intent $T$ into a structured motion specification that supports physically grounded and controllable motion generation. 
Leveraging the zero-shot reasoning capabilities of vision-language models (VLMs), we decouple the process into four specialized agents: a visual-perception agent, a relation-analysis agent, a motion planner, and a clipping refiner.

\subsubsection{Visual-perception Agent: Independent Attribute Extraction}
The visual-perception agent is responsible for extracting interaction-agnostic physical attributes from $I_H$ and $I_O$. 
Crucially, to prevent premature semantic bias, the agent treats the human and object images as independent entities without assuming any shared context.
\begin{itemize}[topsep=0em,leftmargin=1.2em,itemindent=0em,labelsep=1em]
\item For the human image $I_H$, the agent extracts directly observable physical attributes, including coarse body pose, limb and hand configuration, hand occupancy (only when visually evident), gaze direction, and body orientation relative to the camera.
Low-level scene context is also extracted, covering environment type, background characteristics, lighting conditions, support surfaces, and camera framing.
\item For the object image $I_O$, it focuses on descriptive physical properties such as coarse category, geometric shape, approximate physical scale, material appearance, surface texture, structural components, and visible text or markings.
\end{itemize}
The visual-perception agent's output serves as a fact-based, structured set of physical attributes, providing reliable, interpretable inputs for downstream reasoning while avoiding premature semantic coupling.

\subsubsection{Interaction-analysis Agent: Affordance and Intent Modeling}
Building on the physical attributes provided by the visual-perception agent, the relation-analysis agent performs affordance- and intention-aware interaction reasoning by considering the initial prompts $T$.
The primary objective of this agent is to infer plausible human–object interactions. 
Reasoning is carried out along three dimensions: (i) human–object compatibility, assessing whether the human’s pose and hand state can physically support engagement with the object; (ii) affordance-driven interaction hypotheses, such as grasping, holding, operating, wearing, or presenting, which may coexist and are not mutually exclusive; and (iii) intention-level inference, estimating the most likely human goal if an interaction were to occur.

A key constraint is that the initial prompts $T$ act as the highest-priority action anchor. 
Explicit action verbs, hand usage (left, right, or both), hand transfer requirements, placement targets, and dressing or wearing constraints are extracted directly from the prompts. 
The agent is strictly forbidden from introducing actions that are not explicitly stated or clearly implied. 
The result of the relation-analysis agent is a structured interaction specification that includes feasible interaction types, physical and logical constraints, inferred intention, and fine-grained manual action constraints, all tailored to support downstream motion planning.

\subsubsection{Motion Planner: Temporal Decomposition}
The motion planner converts high-level interaction intent inferred from the relation-analysis agent into a temporally explicit action plan suitable for direct use by video generation models. 
Action planning is performed at a strict one-second granularity, with the total video duration limited to a maximum of five seconds. 
The planner decomposes the prompt-anchored interaction into a second-by-second timeline that follows a coherent progression—from preparation, through interaction, to completion or display—while respecting hand assignments, hand transfer requirements, placement targets, and wearing or application constraints.
\reviseH{
In our implementation, the timeline is represented with explicit timestamped intervals (e.g., \texttt{[00:00 - 00:01]}, \texttt{[00:01 - 00:02]}, etc.), while the action labels shown in the teaser figure are simplified for visualization.
}

The generated plan $T_M$ is a concise, structured action timeline with explicit product references and motion-oriented descriptions. 
This representation is intentionally designed to be video-model-ready, enabling controllable and physically plausible HOI video synthesis without introducing ungrounded actions or temporal inconsistencies.

\subsubsection{Clipping Refiner: Semantic-level Physical Audit}
Although the planner generates a temporally structured HOI action timeline, we observe that text-level plans may still implicitly induce clipping artifacts during video generation, especially for wearing or attachment-related interactions. 
To address this issue, we introduce a Clipping Refiner, a lightweight temporal refinement agent that operates purely at the semantic planning level. 
The refiner takes as input the detailed action timeline produced by the planner and audits each action segment for high-risk penetration patterns, such as instantaneous attachment, abrupt large motions, or phrasing that implies passing through the human body or clothing. 
Without introducing new action types or altering the output schema, the refiner minimally rewrites the timeline by redistributing existing actions into staged, physically plausible substeps—e.g., alignment, boundary contact, gradual fitting, and micro-adjustment—while strictly preserving the original action intent specified in the initial prompt. 
This process enforces a temporally coherent, penetration-aware action sequence under fixed-duration and granularity constraints, producing a video-model-ready plan that significantly reduces downstream clipping artifacts without requiring any geometric reasoning or additional visual supervision.

Details of the system prompt are provided in the supplementary materials.

\subsection{Text-driven HOI Video Generation}
\label{sec:method.vdm}

\subsubsection{HOI Video Generation Module}
Following the multi-agent thinking-and-planning process, the resulting execution blueprint is used to condition the video generation module.
Our video generation module $M$ is based on Wan2.1-I2V-14B~\cite{wang2025wan}, which is a DiT-based~\cite{peebles2023dit} video diffusion model using a 3D VAE model~\cite{kingma2013vae} with encoder $Enc$ and decoder $Dec$.
Specifically, given the human image $I_H$ and object image $I_O$, we inject their information into the Wan2.1-I2V backbone from two complementary perspectives: token concatenation and CLIP~\cite{radford2021clip} feature fusion.

Concretely, for the human image $I_H$ and object image $I_O$, their VAE-encoded features are first patchified to obtain token representations $H_H$ and $H_O$. 
These tokens are then concatenated with the video token sequence $H$ along the temporal dimension:
\begin{align}
    H = concat_f(H_{O}, H_{H},H),
\end{align}
where the RoPE~\cite{su2024roformer-rope} is set with offset=-1 and -2 for the token-concat images.
Meanwhile, the CLIP features are extracted from human and object images, projected through an MLP for dimensionality reduction, and injected into the backbone via a cross-attention layer alongside the T5~\cite{raffel2020t5} text embeddings.

\subsubsection{Implicit Text-motion Representation Alignment}

Although the HOI video generation module supports text-to-video synthesis, learning fine-grained motion control from text alone requires large-scale paired text-HOI video data, which are scarce in practice. To address this issue, we adopt a feature-level alignment scheme~\cite{yu2024repa, zhang2025videorepa} that bridges text and motion representations within the video model. 
\reviseH{
The core challenge lies in determining the appropriate source of text–motion aligned features for effective feature alignment. To this end, we leverage the observation that text-to-motion models encode strong language-action alignment priors in their intermediate features. 
We train a video diffusion text-to-motion model $M_{motion}$ on the Wan2.1-1.3B backbone. The model takes a human image $I_H$, an object image $I_O$, and a text prompt $T$ as inputs, the same as in $M$, and produces a motion video $Y_M$. The motion video $Y_M$ is presented by human skeletons and object mask sequences.
The hidden tokens of $Y_M$ encode informative priors on human-object dynamics, which may consistently improve the naturalness of motion and the alignment between text and motion in HOI video generation.
}

Building on the aligned representations, we employ an implicit feature alignment strategy. Given the ground-truth motion sequence $Y_M$, we extract intermediate features $H_{motion}$ from the text-to-motion model $M_{motion}$ at layers $\{0,2,4,6,8,10,12,14\}$ and align them with the corresponding features of the video generation model $M$. For each layer $i$, the hidden states $H^i$ and $H^i_{motion}$ are projected via lightweight three-layer projectors, denoted as $H^i=\text{Proj}(H^i)$ and $H^i_{motion}=\text{Proj}_{motion}(H^i_{motion})$. Instead of directly matching features, we apply the TRD loss~\cite{zhang2025videorepa} to enforce consistency in spatial and temporal relative similarities between $H^i$ and $H^i_{motion}$. Specifically, the spatial relative similarity within a frame is defined as:
\begin{equation}
H_{\text{spatial}}^{d,i,j} = 
\frac{
\mathbf{H}_{\mathbf{v}}^{d,i} \cdot 
\mathbf{H}_{\mathbf{v}}^{d,j}
}{
\|\mathbf{H}_{\mathbf{v}}^{d,i}\| 
\|\mathbf{H}_{\mathbf{v}}^{d,j}\|
},
\end{equation}
and the token relationships across temporal frames, which are used to identify temporal dynamic consistency, are:
\begin{equation}
H_{\text{temp}}^{d,i,j,e} = 
\frac{
\mathbf{H}_{\mathbf{v}}^{d,i} \cdot 
\mathbf{H}_{\mathbf{v}}^{e,j}
}{
\|\mathbf{H}_{\mathbf{v}}^{d,i}\| 
\|\mathbf{H}_{\mathbf{v}}^{e,j}\|
},
\quad
\forall e \in [1,f] \setminus \{d\}, \; 
j \in [1, h w].
\end{equation}
We then align the features of the two models along both spatial and temporal dimensions:
\begin{align}
\mathcal{L}_{\text{TRD}} 
&= \frac{1}{f(h w)^2} 
\sum_{d=1}^{f} \sum_{i,j=1}^{h w} 
\left| 
\mathbf{H}_{\text{spatial}}^{d,i,j} - 
\mathbf{\hat{H}}_{\text{spatial}}^{d,i,j} 
\right| \notag \\
&\quad + 
\frac{1}{f(h w)^2(f-1)} 
\sum_{\substack{d,e=1 \\ e \ne d}}^{f} 
\sum_{i,j=1}^{h w} 
\left| 
\mathbf{H}_{\text{temp}}^{d,i,j,e} - 
\mathbf{\hat{H}}_{\text{temp}}^{d,i,j,e} 
\right|,
\end{align}
where $\mathbf{\hat{H}}_{\text{spatial}}^{d,i,j} $ and $\mathbf{\hat{H}}_{\text{temp}}^{d,i,j,e} $ are the corresponding spatial and temporal relative features extracted from $H_{motion}$.
The final training objective of the model $M$ with flow matching loss~\cite{lipman2022flowmatching} is:
\begin{equation}
    L = L_{FM} + \mu \cdot L_{TRD},
\end{equation}
where $\mu$ is a hyperparameter.
$M_{motion}$ is used during training and removed at inference.  

\subsection{Mixed-source Data Curation and Training Strategy}
\label{sec:method.data}
\subsubsection{Real data and synthetic Data Curation}

\input{figures/dataprocess}

Existing datasets are insufficient to support scalable training of diverse human-object interactions due to sparse and incomplete coverage over the joint distribution of actions and objects, as well as the difficulty of data collection and annotation.
To address this, we construct our training corpus using two complementary strategies as in Fig.~\ref{fig:method.dataproess}: filtering high-quality data from massive online videos and generating paired synthetic data.

\paragraph{Automated HOI data filtering.} We design an automated pipeline to curate high-quality HOI samples from large-scale real-world videos. As shown in Fig.~\ref{fig:method.dataproess}(a), we first apply low-level filters to select videos with sufficient visual quality ($\geq$720P), stable camera motion, and no captions or screen cuts, while labeling human-centric attributes. 
We then perform video-level understanding using InternVL3.5-38B to assess interaction quality, motion dynamics, HOI type, object scale, and occlusions, and to generate detailed HOI action timelines. 
Next, we extract a key frame and construct clean object images via Grounding DINO~\cite{liu2024grounding} and SAM2~\cite{ravi2024sam2}. 
In parallel, we restore clean human poses using FLUX Kontext~\cite{labs2025fluxkontext} and apply pose editing to generate reference images. Following this pipeline, we curate 71K high-quality HOI video clips from 10M online videos.

\paragraph{Synthetic Data Generation.}
Despite the fact that we have an automated HOI data filtering pipeline, it still cannot provide sufficient diversity or balanced motion-object distributions. 
Thus, we supplement our dataset with a proportion of synthesized HOI samples (shown in Fig.~\ref{fig:method.dataproess} (b)). 
Concretely, we first use FLUX to generate images of 85 distinct categories and manually filter them to retain 13,000 high-quality object images. Then, using Gemini’s reference-based image generation (Nano Banana 2), we synthesize HOI images where a human naturally grasps or manipulates the object (wear, open, lie on, ride, etc.). 
Next, we use FLUX again to remove objects so as to get clean reference images. 
Finally, we employ Wan2.2-I2V to animate static HOI images into short motion sequences and regard them as synthetic GT videos. 
We generate 37,000 virtual video clips of diverse motion patterns.
More details could be found in the supplementary materials.

\subsubsection{Training Strategy}
While synthetic data can improve the diversity of object shapes, it also introduces challenges, including degraded visual quality, distorted human structures, and blurred objects, as shown in the following experimental section. To mitigate these issues, we adopt a targeted training strategy in which synthetic data is only applied at high-noise timesteps, specifically the top 30\% of the diffusion process. At these stages, the model primarily learns coarse structural information, allowing it to benefit from the diversity of synthetic samples while minimizing the impact of their low-quality textures.

Specifically, the timestep shift in Wan2.1 (with a shift scale of approximately 5) further biases the training distribution toward the high-noise regime. As a result, although synthetic data is assigned to the top 30\% timesteps, it is effectively utilized primarily at very high-noise timesteps, approximately $t \gtrsim 900$ in a 1000-step schedule.

%% file: figures/pipeline.tex
\begin{figure*}
    \centering
    \includegraphics[width=\linewidth]{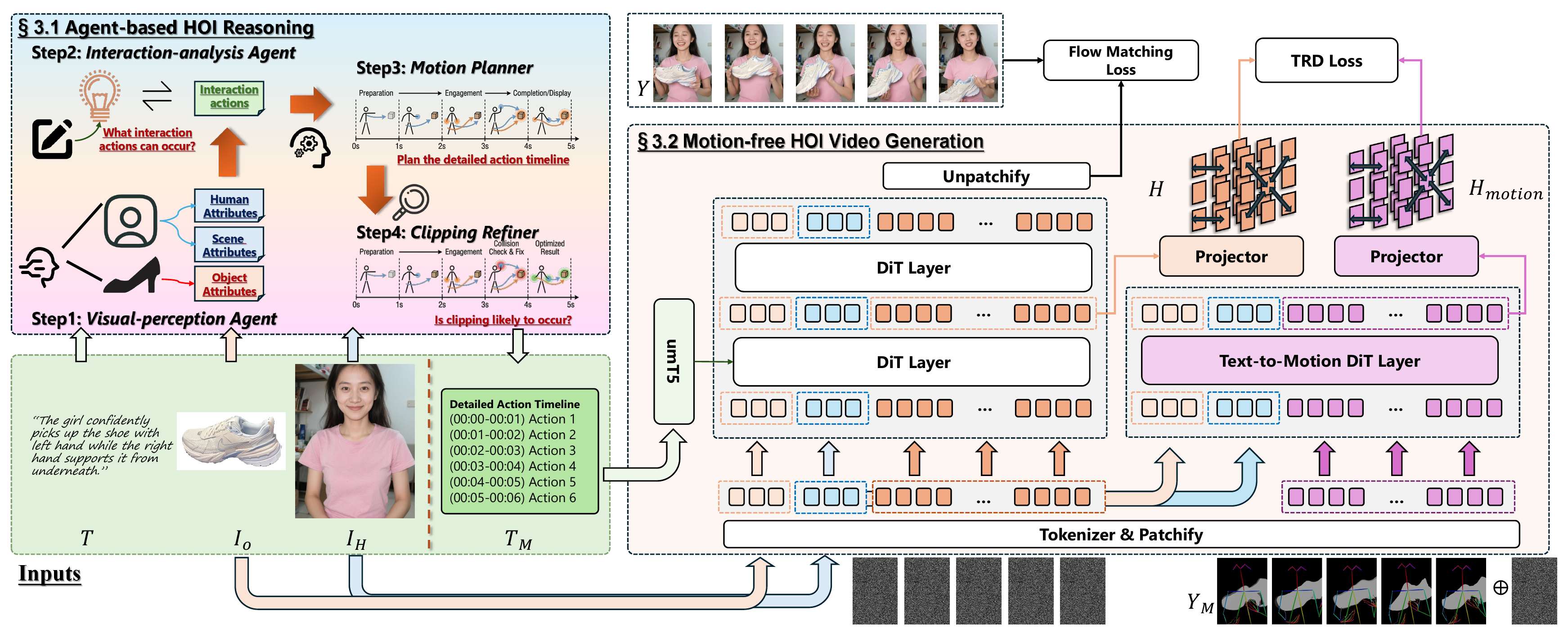}
    \caption{The pipeline for the proposed AgentHOI. Our framework consists of two core components: agent-based HOI reasoning and motion-free HOI video generation. The former employs a multi-agent architecture to infer structured and physically plausible HOI action plans from textual and image inputs, while the latter enhances text controllability by aligning video representations with implicit text-to-motion features during training, enabling robust text-driven HOI video generation without explicit motion inputs at inference time.}
    \label{fig:pipeline}
\end{figure*}

%% file: figures/dataprocess.tex
\begin{figure*}[!t]
    \centering
    \includegraphics[width=\linewidth]{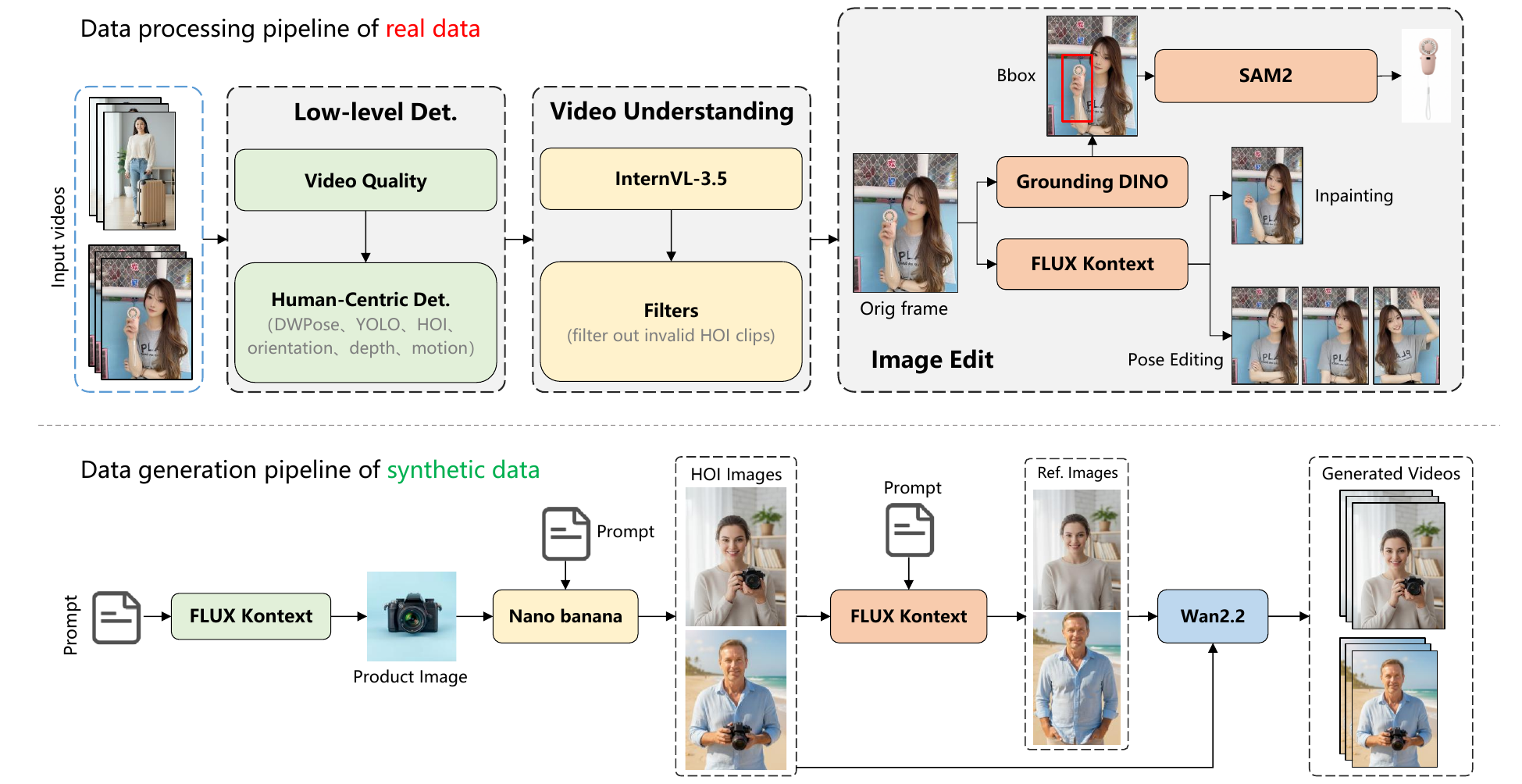}
    \caption{Visualization of real data pipeline and synthetic data pipeline.
    We build an HOI dataset via a hybrid real–synthetic pipeline. For real data, we develop an automated filtering system that mines high-quality HOI clips from large-scale online videos using multi-stage quality control, video understanding, and object/human decomposition, yielding 71K clips from 10M videos. For synthetic data, we generate diverse HOI samples via reference-based human-object synthesis and image-to-video animation, producing 37K additional clips to improve motion and object diversity.
    }
    \label{fig:method.dataproess}
\end{figure*}

%% file: sec/4_expr.tex
\section{Experiments}
\subsection{Experimental Settings}

\subsubsection{Implementation Details}
We use a fixed learning rate of 1e-6 across all stages. 
Training is conducted on 32 GPUs with a total batch size of 16.
Our model is implemented based on WAN2.1-14B.
The total training steps are 9,500 to iterate over all data. 
We use $480$ as our resolution.
For the motion model $M_{motion}$, we use the ground-truth motion video $Y_M$, where human skeletons are extracted using DWPose~\cite{yang2023dwpose}, and objects are segmented with SAM2~\cite{ravi2024sam2, kirillov2023sam}.
We first pre-train $M_{motion}$ on the same dataset using ground-truth motion sequences $Y_M$ for 8000 steps. 
During alignment, $Y_M$ is noised via the diffusion process and fed into $M_{motion}$. 
While predicting denoising noise, the intermediate features $H_{motion}$ capture the motion dynamics of $Y_M$, which are then used for TRD loss with the video generator's features. 
% $M_{motion}$ is used during training and removed at inference.  
For the TRD loss, $\mu$ is set as 0.45.
EMA-VFI~\cite{zhang2023emavfi} is used to interpolate our generated videos to 32 FPS for visual quality, and is not used during the quantitative evaluation phase.

\subsubsection{Dataset}
In total, our training dataset comprises 108,000 samples, consisting of 71,000 real-world data samples and 37,000 synthetic samples.
We evaluate our method primarily under a cross-motion-driven setting, and further validate its generalization on standard self-driven benchmarks.

\textbf{Cross-Motion-Driven Test Set.}
We construct a cross-motion-driven benchmark to explicitly evaluate motion controllability while decoupling motion from appearance. This test set contains 72 samples. 
We extract motion sequences from real human–object interaction videos (sourced from the HOMA benchmark~\cite{huang2025homa}) and manually corresponding text prompts. 
An image generation model (Banana Nano) is used to synthesize the input human and object images.

\textbf{Self-Driven Test Sets.}
We further consider two standard benchmarks where real images are used as inputs.
\begin{itemize}
    \item AnchorCrafter~\cite{xu2024anchorcrafter} benchmark contains four objects and four subjects with relatively simple object categories and interaction patterns. We reorganize this dataset via resampling, resulting in 80 clips.
    \item HOMA Test Set~\cite{huang2025homa}. We adopt the evaluation benchmark introduced in HOMA, which features a broader range of object categories than AnchorCraft. The dataset includes 100 clips with more diverse object appearances and more complex human–object interactions.
\end{itemize}

We report both quantitative and qualitative results on the cross-motion-driven setting, and provide additional evaluations on self-driven benchmarks to assess generalization under more diverse conditions.

\subsubsection{Comparison Methods}
We compare our framework with the following methods:
AnchorCrafter~\cite{xu2024anchorcrafter} is a human-object interaction human animation approach that requires fine-tuning on the target object, and is conducted only on the AnchorCrafter test set.
AnimateX~\cite{AnimateX2025} uses implicit and explicit motion transformation to align the target human identity.
Unianimate-DiT~\cite{wang2025unianimatedit} generates a human video from a person image and a sequence of poses based on the Wan-14B model.
VACE-14B~\cite{jiang2025vace} is a general video editing framework based on the 14B video model Wan~\cite{wang2025wan}, which supports text control.
HUMO~\cite{chen2025humo} supports generating videos conditioned on multiple reference images.
HOMA~\cite{huang2025homa} proposes a weak motion guidance for HOI human video generation, which still needs pose alignment during inference.

The set of compared baselines varies across test sets based on method applicability. AnchorCrafter requires per-object fine-tuning and is therefore evaluated only on its own test set. Animate-X is specifically designed for cross-motion-driven (cross-identity pose-driven) generation and is therefore evaluated only on the cross-motion-driven test set; given its limited overall capability on HOI scenarios, we do not include it on the self-driven HOMA and AnchorCrafter test sets, where the stronger DiT-based UniAnimate-DiT already serves as the pose-driven representative. The remaining baselines (UniAnimate-DiT, VACE-14B, HuMo, and HOMA) are evaluated on all three test sets.

\subsubsection{Metrics}
We evaluate our method using several established metrics. 
Object DINO (\textbf{Obj-DINO})~\cite{xu2024anchorcrafter} is used to evaluate the object consistency by computing the DINO~\cite{oquab2023dinov2} similarity of the objects between the input object image and the generated videos within the segmentation.
Dynamic degree (\textbf{DD}), aesthetic quality (\textbf{AQ}), and motion smoothness (\textbf{MS}) extracted from Vbench~\cite{huang2024vbench} are employed to measure general video quality.
TVA~\cite{liu2025videoalign} is utilized to measure the alignment between the text prompt and the videos~\cite{chen2025humo}, and is not computed for pose-driven methods.
We further use InternVL3-38B~\cite{chen2024internvl, zhu2025internvl3} to identify object consistency ($IVL_{O}$), human consistency ($IVL_{H}$), and interaction ($IVL_{I}$) quality with structured prompts. Details of the metrics are provided in the supplementary materials.

\subsection{Comparison Results}
\label{expr:main_quan}
\input{tables/main}

\subsubsection{Results of Cross-Motion-Driven Test Set.}
As shown in Table~\ref{tab:main}, our method consistently outperforms all baselines across key metrics. 
In particular, we achieve the highest Obj-DINO score, demonstrating superior object appearance preservation. 
We also obtain the best DD, AES, and MS, indicating strong overall video quality with smooth and coherent motion. 
Benefiting from the agent-based HOI reasoning and planning module, our method obtains the best TVA, showing the strongest adherence of motion dynamics to the input text instructions. 
Finally, the VLM-based evaluation with InternVL3 shows that our results achieve the highest object consistency, human consistency, and interaction quality, matching the best scores and substantially surpassing other methods, validating the effectiveness of our agent-based HOI reasoning and motion-free generation framework.

\input{figures/main_expr}

As shown in Fig.~\ref{fig:expr.main_expr}, we qualitatively compare our method with recent HOI/human-video baselines on two representative prompts (bimanual vs. single-hand interactions). Our method best preserves both human identity and object appearance, while producing physically plausible hand-object contact and instruction-faithful motion dynamics.

In contrast, Animate-X generally maintains the reference appearance but exhibits weak interaction expressiveness, where the injected object is almost motionless. UniAnimate-DiT can execute the coarse motion pattern, but it frequently suffers from object absence and fails to perform fine-grained hand actions. Due to the difficulty of retargeting, HOMA produces incorrect hand synthesis and renders the object with incorrect textures, leading to implausible interaction results. VACE-14B tends to introduce severe object deformation or incorrect scaling, which degrades appearance fidelity. HUMO exhibits severe object artifacts such as splitting and deformation. Overall, all other methods fail to satisfy the text prompts for fine-grained hand-object interactions, whereas our approach consistently follows the action instructions while maintaining stable appearances and realistic contact dynamics, validating the effectiveness of our agent-based HOI reasoning and motion-free generation framework.

\subsubsection{Results of Self-Driven Test Set.}
\input{figures/main_real}

\input{tables/real_testset}
To further verify that the improvements are not specific to our cross-motion-driven benchmark, we additionally evaluate on two self-driven test sets with real input images: the HOMA test set~\cite{huang2025homa} and the AnchorCrafter test set~\cite{xu2024anchorcrafter}. Since these benchmarks are originally motion-driven, we manually annotate text prompts for each clip so that all text-driven baselines and our method receive comparable inputs. Quantitative results are reported in Table~\ref{tab:real_testset_split}, and qualitative comparisons are shown in Fig.~\ref{fig:expr.main_homa} and Fig.~\ref{fig:expr.main_AC}.

On the HOMA test set, our method achieves the best DD, MS, TVA, and InternVL-based interaction scores, and substantially outperforms the text-driven baselines VACE-14B and HUMO on Obj-DINO, indicating better object appearance preservation under real input images. The motion-driven HOMA baseline obtains a slightly higher Obj-DINO score, which is largely attributed to its DINO-feature injection during generation; nevertheless, our method still surpasses it on the InternVL-based interaction quality and motion-text alignment, showing that text-conditioned reasoning produces more faithful interaction dynamics. On the AnchorCrafter test set, our method consistently leads on Obj-DINO, TVA, and all three InternVL-based metrics, while remaining competitive on AES and MS. The qualitative comparisons in Fig.~\ref{fig:expr.main_homa} and Fig.~\ref{fig:expr.main_AC} further show that our framework preserves both the human and object appearances under real-image inputs, follows multi-step instructions, and maintains stable hand-object contact, which validates that the gains observed on the cross-motion-driven benchmark generalize to real-image self-driven settings as well.

\subsection{Ablation studies and discussions}
\subsubsection{Analysis of agent}

\input{figures/abla_agent}

We ablate the HOI reasoning and planning module by changing how the motion description $T_M$ is produced from $(I_H, I_O, T)$, while keeping the video generator and all inference settings fixed. Fig.~\ref{fig:abla_agent} provides representative visual comparisons.

\paragraph{No-Agent} directly uses the input without structured reasoning or temporal planning, forcing the generator to rely on implicit priors for interaction details. As shown in the top row of Fig.~\ref{fig:abla_agent}, this often leads to unstable object scale and weaker interaction cues, whereas our full system maintains more plausible object size and presentation. Quantitatively, removing agents causes consistent drops in TVA and InternVL-obj, indicating weaker motion-text alignment and reduced object consistency (Table~\ref{tab:main}).

\paragraph{Single-Agent} uses a monolithic VLM to produce $T_M$ in one pass. Although it introduces explicit intent, entangled perception and planning can induce ungrounded details. In the middle row of Fig.~\ref{fig:abla_agent}, the single-agent variant tends to hallucinate extra objects or duplicate the target (e.g., multiple shoes with inconsistent appearance), while our method preserves the correct object identity and stable appearance across frames. This is also reflected by lower Obj-DINO and reduced InternVL-inter compared with the full system (Table~\ref{tab:main}).

\paragraph{No Clipping Refiner.} We keep the three-agent decomposition but remove the Clipping Refiner, using the raw planned timeline as $T_M$. Without this penetration-aware semantic refinement, attachment-related interactions become less stable (bottom row in Fig.~\ref{fig:abla_agent}). Correspondingly, TVA and InternVL-inter decrease, suggesting that the refiner improves both instruction-faithful motion dynamics and interaction quality (Table~\ref{tab:main}).

\subsubsection{Validation of Implicit Text-Motion Representation Alignment}
We validate the proposed implicit alignment by removing it during training and by an inference-time baseline that directly injects explicit motion cues. As shown in Table~\ref{tab:main} and Fig.~\ref{fig:ablation_alignment}, disabling alignment degrades both motion-text faithfulness (TVA) and interaction quality, and often leads to less coherent motion progression and unstable object handling. 
Directly adding motion guidance at inference provides limited benefits and can introduce additional instability, as the injected motion prior is generated by the text-to-motion model. 
In contrast, our implicit feature alignment distills text-to-motion priors into the video model during training using ground-truth motion, leading to more stable object appearance and more natural, instruction-faithful interaction dynamics, while introducing no additional overhead at inference time.

\input{figures/abla_implicit}

\input{figures/abla_trd}
\rewriteH{
\paragraph{TRD Alignment-Layer Selection.} We further study which intermediate layers of $M_{motion}$ should be used for TRD alignment. As shown in Fig.~\ref{fig:abla_trd}, aligning with low-layer features captures motion structure while staying disentangled from fine-grained appearance details, whereas aligning with high-layer features tends to propagate over-specific patterns and errors from $M_{motion}$ into the video generator. Quantitatively, the ``w/ high layers selection'' variant in Table~\ref{tab:main} drops Obj-DINO from 0.510 to 0.481, DD from 0.986 to 0.903, and TVA from 0.643 to 0.520, while InternVL-based scores also decrease. These results support our default choice of aligning lower intermediate layers, which provides motion-structure guidance without overconstraining the appearance and motion priors of the video backbone.
}

\subsubsection{Validation of Mixed-Source Data}
\input{figures/abla_data}
We examine the effect of incorporating synthetic data into the training set.
As shown in Table~\ref{tab:main} and Fig.~\ref{fig:ablation_data}, the addition of synthetic samples significantly improves the model’s understanding of object information and interactions.
This improvement is largely attributed to the limited diversity of real-world data, which cannot fully cover the distribution of possible object types and interactions.
By supplementing the dataset with high-quality synthetic samples, we effectively alleviate this limitation, enhancing both generalization and action fidelity.

\input{figures/abla_onlysyndata}

\rewriteH{
\paragraph{Only Synthetic Data.} To further isolate the contribution of synthetic samples, we additionally train a variant using only the synthetic subset. As shown in Table~\ref{tab:main} and Fig.~\ref{fig:abla_onlysyndata}, the only-synthetic variant improves over the no-synthetic baseline (Obj-DINO 0.371 $\to$ 0.450, DD 0.797 $\to$ 0.902, TVA 0.517 $\to$ 0.526, InternVL-obj 0.870 $\to$ 0.919), confirming that synthetic data does provide informative supervision for object appearance and interaction patterns. However, it still lags behind the full system (Obj-DINO 0.510, DD 0.986, TVA 0.643, InternVL-obj 0.970), with visibly more residual texture artifacts and less stable interaction dynamics. This indicates that the strong performance of our final model arises from the combination of real-world realism and synthetic diversity rather than synthetic data alone.
}

\paragraph{Analysis of Training Strategies}
Since synthetic samples may contain appearance artifacts such as blurred textures or distorted human figures, we use them only at high diffusion noise steps, where the model focuses on coarse structure and motion rather than fine textures. As shown in Table~\ref{tab:main} and Fig.~\ref{fig:ablation_data}, this strategy retains most of the benefits of synthetic diversity while reducing sensitivity to synthetic appearance noise. Compared with naive mixed-source training, it produces fewer texture artifacts and more stable interaction dynamics.

\subsubsection{Analysis of Different VLMs}
\input{tables/ablation_vlm}
\input{figures/abla_diffvlm}
In this section, we analyze the impact of different visual language models (VLMs) on agent performance. Specifically, we compare the capabilities of InternVL-3.5-38B, Gemini-Pro (Gemini-2.5-Pro), Gemini-Pro-T (text-only variant) and Qwen-VL-7B, as presented in Table~\ref{tab:ablation_vlms} and Fig.~\ref{fig:abla_diffvlm}.

Our results demonstrate that model performance is primarily influenced by the alignment between the reasoning outputs and the video generation process, rather than by the absolute capability of the VLM itself. While Gemini-Pro is generally considered to be more capable than the other compared VLMs, it tends to produce longer and more detailed timelines with excessive descriptive content, increasing the reliance on text and diminishing the consistency with the reference image. In contrast, InternVL generates more concise action phrases, which better preserve object consistency across frames and ensure smoother interaction with the visual input. In contrast, the smaller-scale Qwen-VL-7B tends to produce overly simplistic plans, often consisting of only a few words, which leads to inferior video generation results compared to the stronger VLMs.

To isolate the effects of structured visual reasoning, we introduce a baseline using the text-only version of Gemini-Pro as Gemini-Pro-T. This comparison reveals that visual reasoning significantly improves interaction consistency, offering a clear advantage over simple prompt expansion.

\rewriteH{
\subsection{Further Comparisons of our Agent Pipeline}
\input{tables/agent_fairness}
\input{figures/abla_faircomp}
To further isolate the contribution of our multi-agent HOI reasoning module from the rest of the framework, we additionally attach the same agent pipeline to two strong text-driven baselines, VACE-14B~\cite{jiang2025vace} and HUMO~\cite{chen2025humo}, and feed the resulting structured motion timeline as their text input. Quantitative results are reported in Table~\ref{tab:agent_fairness}, and qualitative comparisons are shown in Fig.~\ref{fig:abla_faircomp}.
}

Equipping VACE-14B and HUMO with our agent pipeline brings clear improvements over their original text-driven counterparts: Obj-DINO improves from 0.328 to 0.505 for VACE-14B and from 0.390 to 0.493 for HUMO, while TVA, InternVL-obj, and InternVL-inter also increase consistently. This confirms that the structured perception, relation-analysis, and motion-planning agents provide a generally beneficial prompt-refinement effect that is not specific to our video backbone. Even with this enhancement, both ``+agent'' baselines still fall behind our full system on Obj-DINO, MS, TVA, and InternVL-inter, indicating that the implicit text-motion representation alignment and the mixed-source data pipeline contribute complementary improvements that go beyond what better prompts alone can achieve.

\subsection{More Results}
\input{figures/app}
As illustrated in Fig.~\ref{fig:app}, we present additional qualitative results covering challenging object categories such as small objects and deformable objects (e.g., clothing), which are typically difficult for pose-driven methods to handle.
We use Z-Image~\cite{cai2025zimage} to generate input images to avoid copyright and redistribution issues.

\paragraph{Small/Large Objects}
We further present interactions with objects at very different scales, ranging from small objects (e.g., operating a robot vacuum) to large objects (e.g., presenting a car) in Fig.~\ref{fig:app}. These cases stress precise hand contact and stable object rendering for small targets, while requiring coherent full-body motion and consistent spatial relationships for large objects. Our method preserves object appearance and maintains plausible interaction dynamics across frames under both scale extremes.

\paragraph{Wearing}
We show body-attached interactions such as wearing a jacket and carrying a backpack (Fig.~\ref{fig:app}). These cases are challenging due to close contact, heavy occlusion, and non-rigid deformations. Our method follows the wearing instructions with a coherent step-by-step progression and maintains stable appearance of the attached items throughout the sequence.

\paragraph{Multi-action}
We include multi-stage prompts involving consecutive actions, such as pick-and-place followed by re-grasping and left-to-right hand transfer (Fig.~\ref{fig:app}). Compared with single-action prompts, these sequences require correct action ordering, smooth transitions, and precise hand assignment. Our method executes the specified stages consistently while keeping the manipulated object identity and appearance stable across frames.

\section{User study}
We conduct a user study using A/B comparisons on all three test sets. For each test set, our method is paired against every baseline that appears in the corresponding quantitative table. Each study involves 30 participants, and each participant completes 40 trials per test set. In each trial, participants are shown the input human image, object image, and text prompt, along with two videos displayed side by side in random left/right order, and are asked to select the better one based on five criteria: human consistency, object consistency, text adherence, interaction naturalness, and overall video quality. Trial order is randomized, and short breaks are provided to mitigate fatigue.

For each baseline and criterion, we report the percentage of trials in which our method is preferred over the baseline. As shown in Table~\ref{tab:userstudy}, our method achieves substantial preference margins on the cross-motion-driven test set, exceeding 89\% against pose-driven baselines (Animate-X, UniAnimate-DiT, HOMA) and ranging from 73\% to 88\% against text-driven baselines (VACE-14B, HuMo). On the self-driven HOMA and AnchorCrafter test sets, our method is also preferred over all baselines on every criterion, with the narrowest yet still positive margin appearing against the motion-driven HOMA baseline on its own test set (60\%--63\%). We show the user-study interface in the supplementary material.

\input{tables/userstudy}

\section{Limitations}
\input{figures/limit}
Due to resolution constraints, text rendered within some generated videos may appear less clear, as exemplified in Fig.~\ref{fig:limit}. Extending the proposed framework to support higher-resolution generation (e.g., 720p) represents a natural direction for future work and may help alleviate this limitation.

AgentHOI also does not enforce strict timing control, and VLM hallucinations in the multi-agent pipeline can occasionally lead to erroneous action timelines (e.g., misidentified objects or missing actions) that propagate into the generated video.

%% file: tables/main.tex
\begin{table}
  \centering
  \caption{Quantitative results of our method compared with SOTAs. }
    \label{tab:main}
  % \vspace{-4mm}
  \resizebox{0.8\linewidth}{!}{
  \begin{tabular}{@{}lcccccccccc@{}} 
    \toprule
                                                              & Obj.-DINO$\uparrow$    &DD$\uparrow$           &AES$\uparrow$            & MS$\uparrow$                  & TVA$\uparrow$           & InternVL$_{O}\uparrow$   & InternVL$_{H}\uparrow$  & InternVL$_{I}\uparrow$   \\
    \midrule
    Animate-X                                                 &  0.2739       &  0.1666       & 0.5710         & 0.9900              &  -               & 0.9708         & 0.9888           &  0.9666    \\
    UniAnimate-DiT                                            & 0.2426         &  0.5000       & 0.5457         & 0.9917             & -            & 0.7666         & 0.9722           &  0.6777    \\
    VACE-14B                                                  & 0.3279         &  0.7916       & 0.5672         & 0.9853             &0.4766        & 0.8555         & 0.9722           &  0.8736    \\
    HUMO                                                      & 0.3897         &  0.9307       & 0.5237         & 0.9925             &0.3692          & 0.8041         & 0.9277           &  0.8055    \\
    HOMA                                         & 0.2788         &  0.8333       & 0.5212         & 0.9897             & -            & 0.7361         & 0.9750           &  0.6388    \\
    Ours                                                      & \textbf{0.5107}&  \textbf{0.9861} & \textbf{0.5818} & \textbf{0.9957}                &\textbf{0.6436} & \textbf{0.9708}      & \textbf{0.9888}     &  \textbf{0.9666}    \\
    \midrule
    w/o agent                                            & 0.4905         &  0.9303       & 0.5548         & 0.9835             & 0.5789         & 0.8913         & 0.9612           &  0.9666    \\
    w/ single agent                                      & 0.4818         &  0.9215       & 0.5390         & 0.9833             & 0.5881         & 0.9112         & 0.9713           &  0.9000    \\
    w/o clipping agent                                   & 0.5047         &  0.9771       & 0.5712         & 0.9977             & 0.6169         & 0.9523         & 0.9888           &  0.9569    \\
    \midrule
    w/o implicit alignment                                & 0.4548        & 0.8472        & 0.5669         & 0.9822            & 0.6098        &   0.9500        & 0.9861            & 0.9472 \\
    w/  explicit injection                                 & 0.4355        &  0.875          & 0.5582         & 0.9832             & 0.5622     &   0.9736        &  0.9875           & 0.9569  \\
   w/ high layers selection                               & 0.481         &0.903          & 0.591           & 0.983            & 0.520         & 0.962           & 0.987              & 0.950  \\
    \midrule
    w/o syndata                                            & 0.3718      &   0.797          &  0.5678            &   0.9844                        &    0.5176       &    0.8708         & 0.9819          & 0.9000       \\
    w/o training strategy                                  & 0.4748       &  0.881          &  0.5027            &  0.9807                         & 0.4469          &    0.8875        & 0.9702           & 0.8861      \\
    Only syndata                                  & 0.450       &  0.902          &  0.577            &  0.982                         & 0.526          &    0.919        & 0.975           & 0.933      \\
    \bottomrule
  \end{tabular}
  }
\end{table}

%% file: figures/main_expr.tex
\begin{figure*}[htbp]
    \centering
    \includegraphics[width=\linewidth]{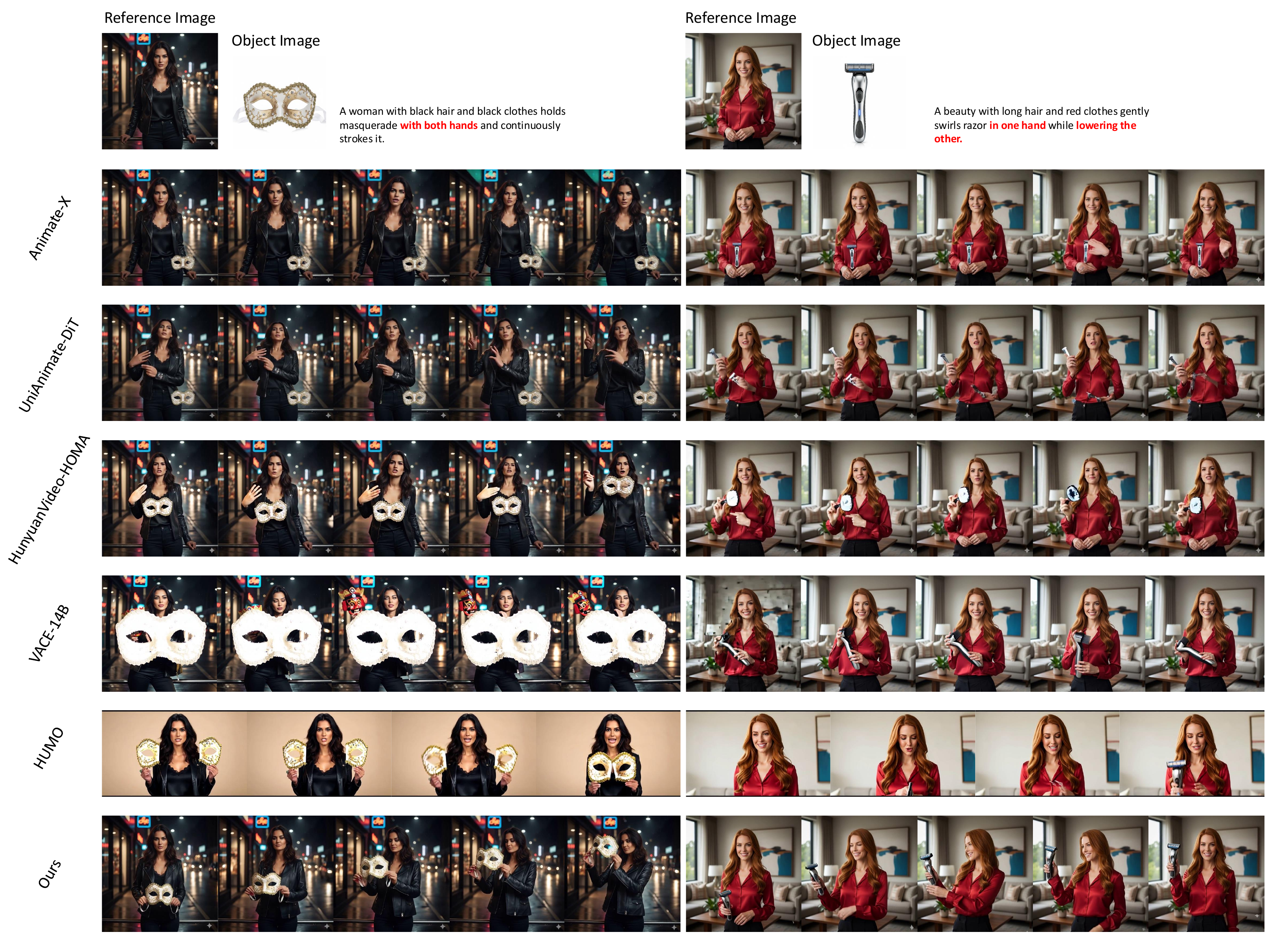}
    % \vspace{-7mm}
    \caption{Comparison with SOTAs. Video results can be found in the supplementary materials.}
    \label{fig:expr.main_expr}
\end{figure*}

%% file: figures/main_real.tex
\begin{figure}[htbp]
    \centering
    
    \includegraphics[width=\linewidth]{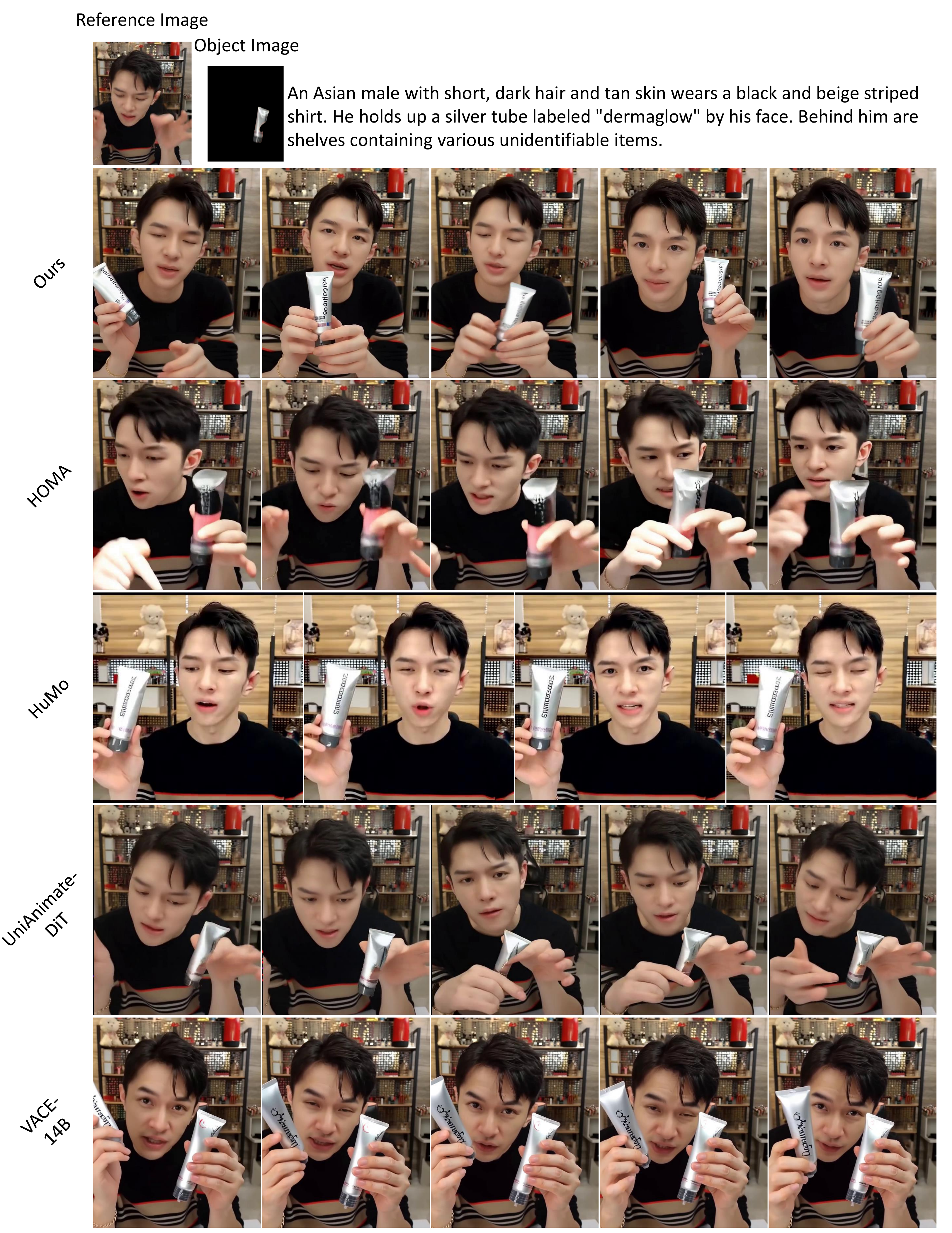}
    % \vspace{-7mm}
    \caption{Comparison with SOTAs on HOMA test set. Video results can be found in the supplementary materials.}
    \label{fig:expr.main_homa}
\end{figure}

\begin{figure}[htbp]
    \centering
    
    \includegraphics[width=0.7\linewidth]{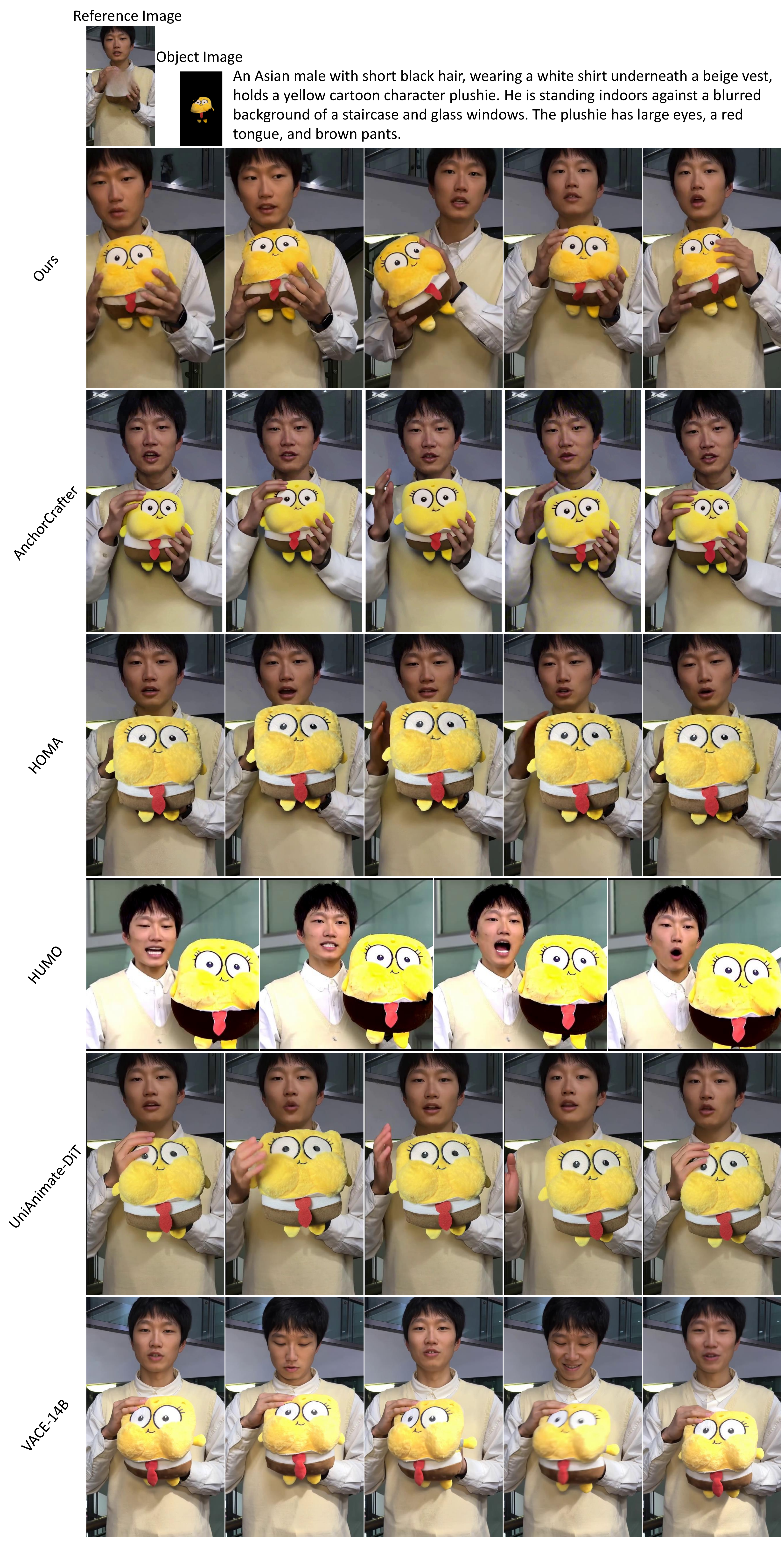}
    % \vspace{-7mm}
    \caption{Comparison with SOTAs on AnchorCrafter test set. Video results can be found in the supplementary materials.}
    \label{fig:expr.main_AC}
\end{figure}

%% file: tables/real_testset.tex
\begin{table}[t]
  \centering
  \caption{Quantitative results of our method compared with SOTAs on HOMA and AnchorCrafter testset.}
  \label{tab:real_testset_split}
  \begin{subtable}[t]{0.8\linewidth}
    \centering
    \caption{Results on HOMA test set.}
    \label{tab:homa_split}
    \resizebox{\linewidth}{!}{
    \begin{tabular}{@{}lcccccccc@{}}
      \toprule
       & Obj.-DINO↑ & DD↑ & AES↑ & MS↑ & TVA↑  & InternVL$_{O}\uparrow$   & InternVL$_{H}\uparrow$  & InternVL$_{I}\uparrow$   \\
      \midrule
      Unianimate-Dit & 0.471 & 0.870 & 0.518 & 0.990 & - & 0.800 & 0.972 & 0.818 \\
      VACE-14B & 0.479 & 0.816 & 0.535 & 0.990 & 1.115 & 0.880 & 0.997 & 0.842 \\
      HUMO     & 0.480 & 0.348 & 0.510 & 0.994 & 0.595 & 0.758 & 0.926 & 0.769 \\
      HOMA     & 0.655 & 0.902 & 0.586 & 0.981 & -     & 0.887 & 0.983 & 0.885 \\
      Ours     & \textbf{0.612} & \textbf{0.906} & \textbf{0.525} & \textbf{0.991} & \textbf{1.214} & \textbf{0.901} & \textbf{0.995} & \textbf{0.902} \\
      \bottomrule
    \end{tabular}
    }
  \end{subtable}

  \vspace{4mm} 

  \begin{subtable}[t]{0.8\linewidth}
    \centering
    \caption{Results on AnchorCrafter test set.}
    \label{tab:anchor_split}
    \resizebox{\linewidth}{!}{
    \begin{tabular}{@{}lcccccccc@{}}
      \toprule
      & Obj.-DINO↑ & DD↑ & AES↑ & MS↑ & TVA↑  & InternVL$_{O}\uparrow$   & InternVL$_{H}\uparrow$  & InternVL$_{I}\uparrow$   \\
      \midrule
      AnchorCrafter & 0.361 & 0.846 & 0.503 & 0.991 & -     & 0.725 & 0.800 & 0.748 \\
      Unianimate-Dit & 0.371 & 0.940 & 0.523 & 0.992 & - & 0.728 & 0.968 & 0.748 \\
      VACE-14B      & 0.397 & 0.915 & \textbf{0.539} & \textbf{0.994} & 0.649 & 0.745 & 0.791 & 0.733 \\
      HUMO          & 0.451 & 0.366 & 0.530 & \textbf{0.994} & 0.701 & 0.849 & 0.915 & 0.776 \\
      HOMA          & 0.336 & \textbf{1.000} & 0.515 & 0.992 & -     & 0.703 & 0.804 & 0.742 \\
      Ours          & \textbf{0.529} & 0.971 & 0.500 & 0.987 & \textbf{1.283} & \textbf{0.952} & \textbf{0.997} & \textbf{0.900} \\
      \bottomrule
    \end{tabular}
    }
  \end{subtable}
\end{table}

%% file: figures/abla_agent.tex
\begin{figure}
    \centering
    \includegraphics[width=1.\linewidth]{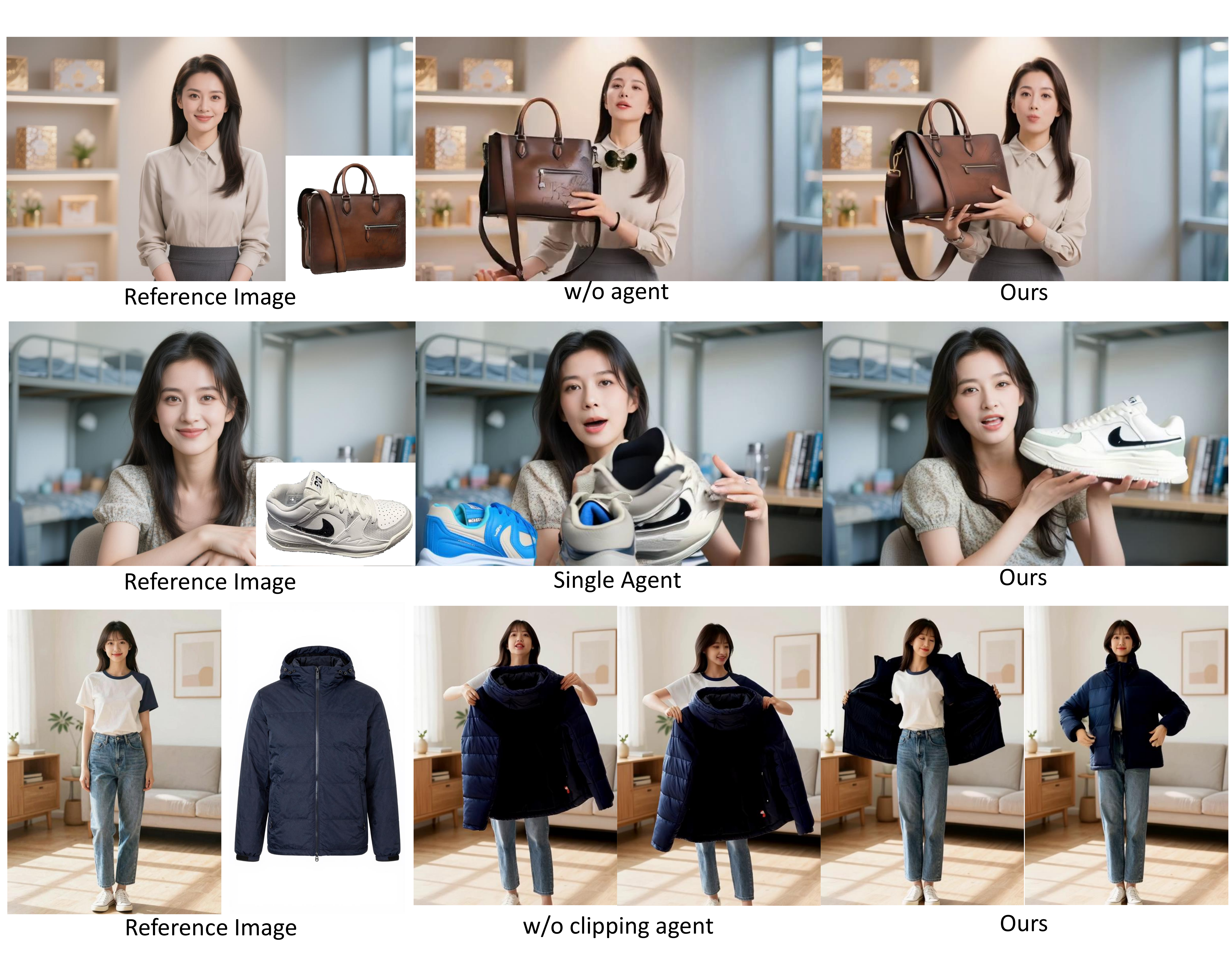}
    \caption{Ablation on the agent-based HOI reasoning and planning module. More analysis of the optimized prompt is shown in the supplementary material.}
\label{fig:abla_agent}
\end{figure}

%% file: figures/abla_implicit.tex
\begin{figure}
    \centering
    \includegraphics[width=1.\linewidth]{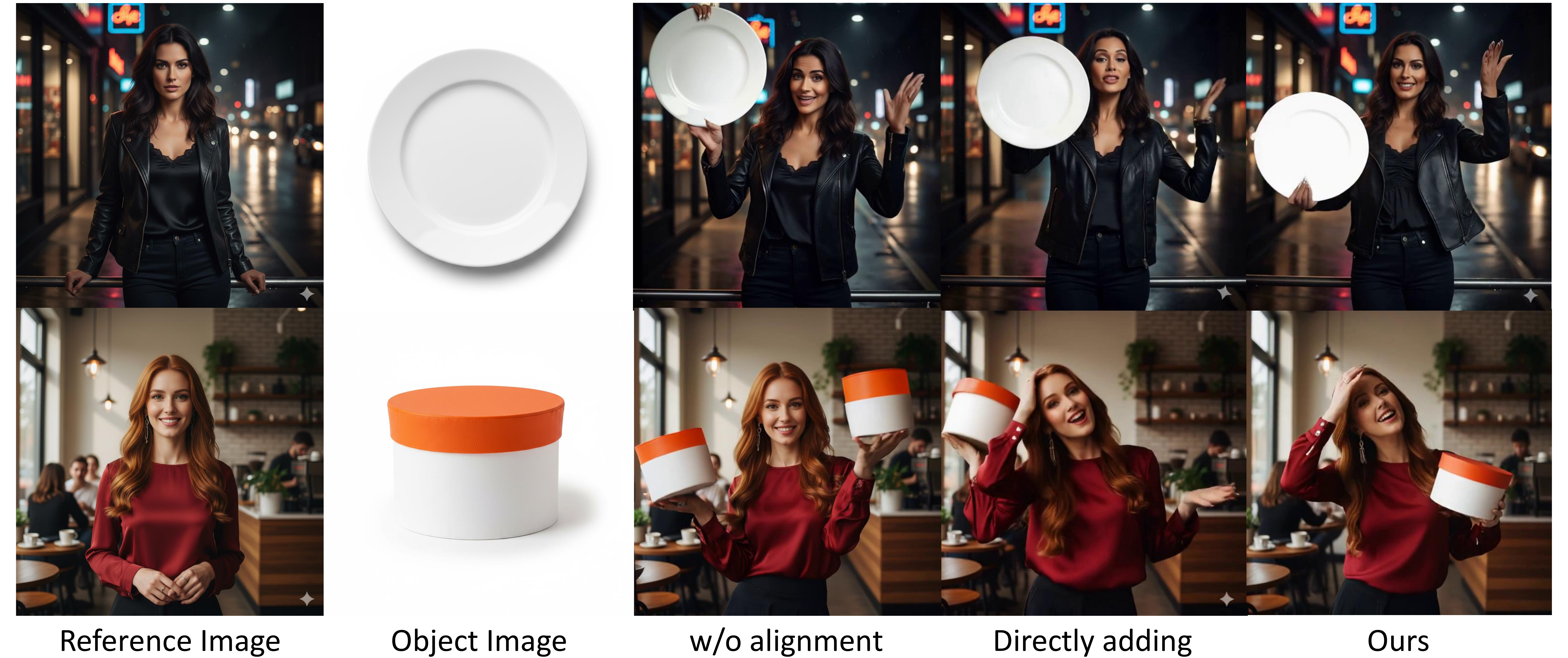}
    \caption{Ablation on implicit text--motion representation alignment.}
\label{fig:ablation_alignment}

\end{figure}

%% file: figures/abla_trd.tex
\begin{figure}
    \centering
    \includegraphics[width=1.\linewidth]{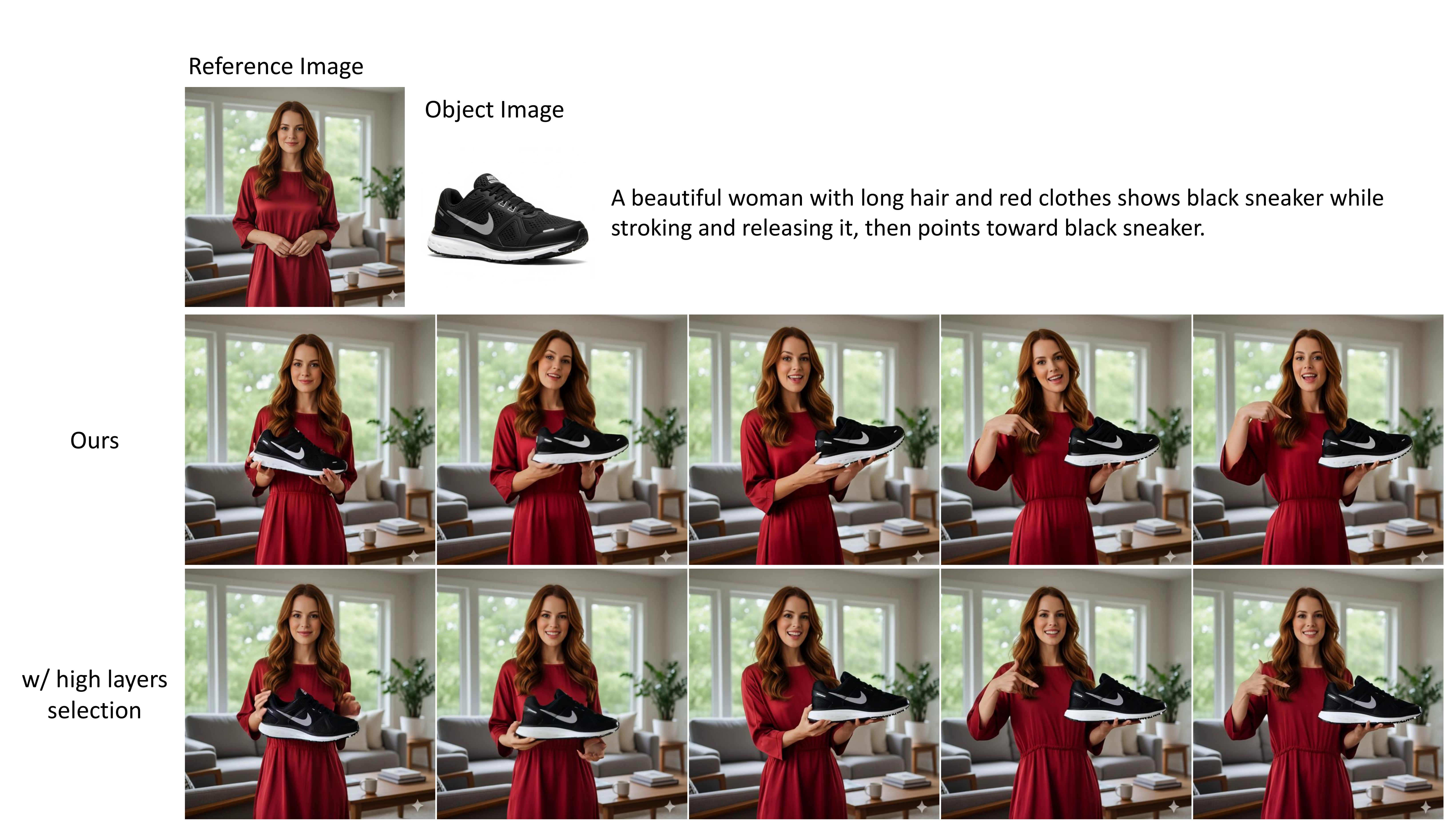}
    \caption{Effect of TRD alignment-layer selection: low-layer features (ours) vs.\ high-layer features.}
\label{fig:abla_trd}

\end{figure}

%% file: figures/abla_data.tex
\begin{figure}
    \centering
    \includegraphics[width=1.\linewidth]{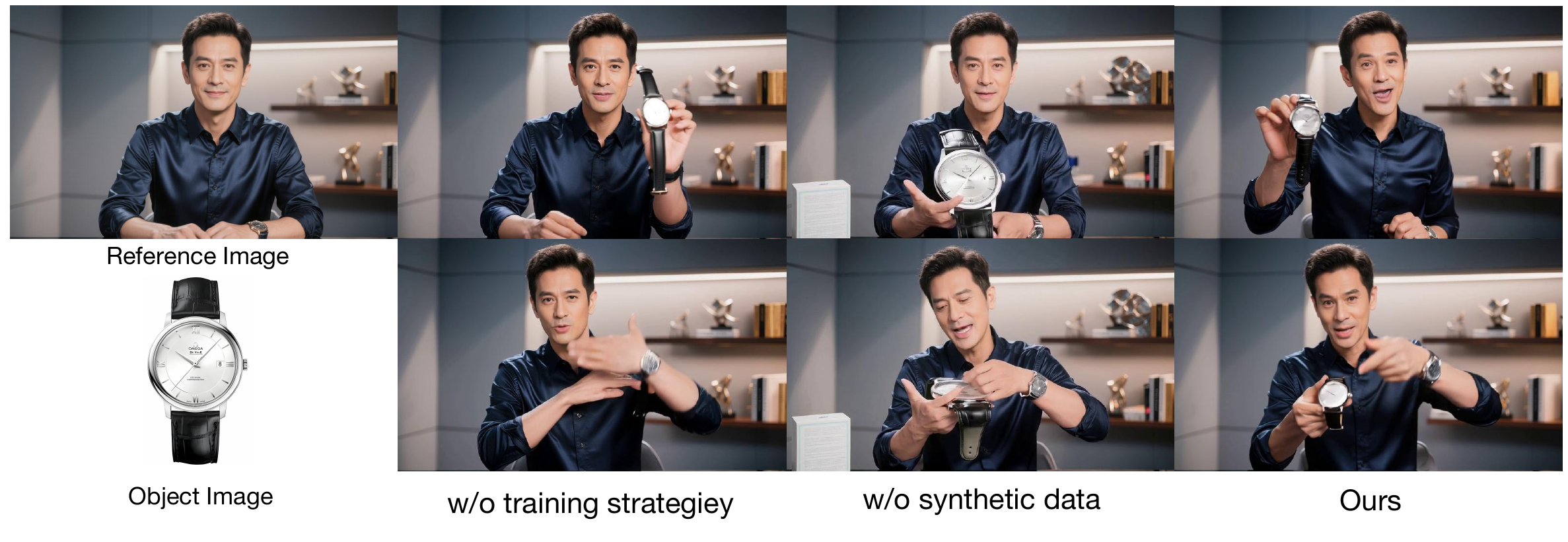}
    \caption{Ablation on mixed-source data and the high-noise synthetic training strategy.}
\label{fig:ablation_data}

\end{figure}

%% file: figures/abla_onlysyndata.tex
\begin{figure}
    \centering
    \includegraphics[width=0.8\linewidth]{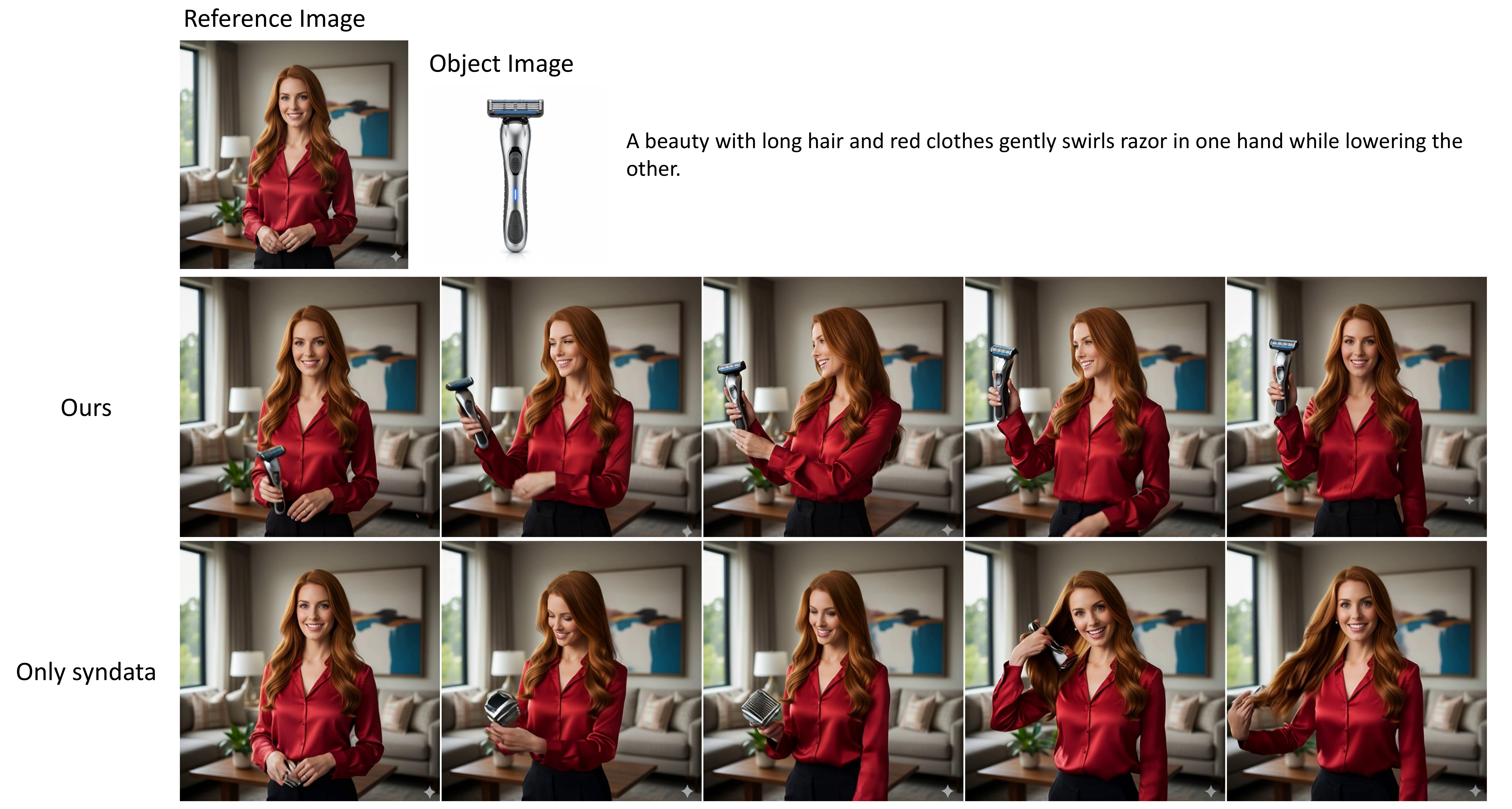}
    \caption{Results of training with only synthetic data, compared to the full mixed-source training.}
\label{fig:abla_onlysyndata}
\end{figure}

%% file: tables/ablation_vlm.tex
\begin{table}
  \centering
  \caption{Quantitative results with different VLMs.}
    \label{tab:ablation_vlms}
  \resizebox{1\linewidth}{!}{
  \begin{tabular}{@{}lcccccccccc@{}} 
    \toprule
                  & Obj.-DINO↑    &DD↑           &AES↑            & MS↑                  & TVA↑           & InternVL$_{O}\uparrow$   & InternVL$_{H}\uparrow$  & InternVL$_{I}\uparrow$   \\
    \midrule
    InternVL       & 0.510         &  0.986       & 0.581         & 0.995            & 0.643        & 0.970         & 0.988           &  0.966    \\
    Gemini-pro             & 0.506         &  0.888       & 0.564         & 0.984             & 0.600          & 0.973         & 0.986           &  0.963    \\
    Gemini-pro—T & 0.478         &  0.902       & 0.563         & 0.979             & 0.552            & 0.920         & 0.977           &  0.901    \\
    Qwen-VL-7B             & 0.498         &  0.861         & 0.556      & 0.983             &0.611         & 0.956        & 0.988           &  0.938    \\
    \bottomrule
  \end{tabular}
  }
\end{table}

%% file: figures/abla_diffvlm.tex
\begin{figure}
    \centering
    \includegraphics[width=1.\linewidth]{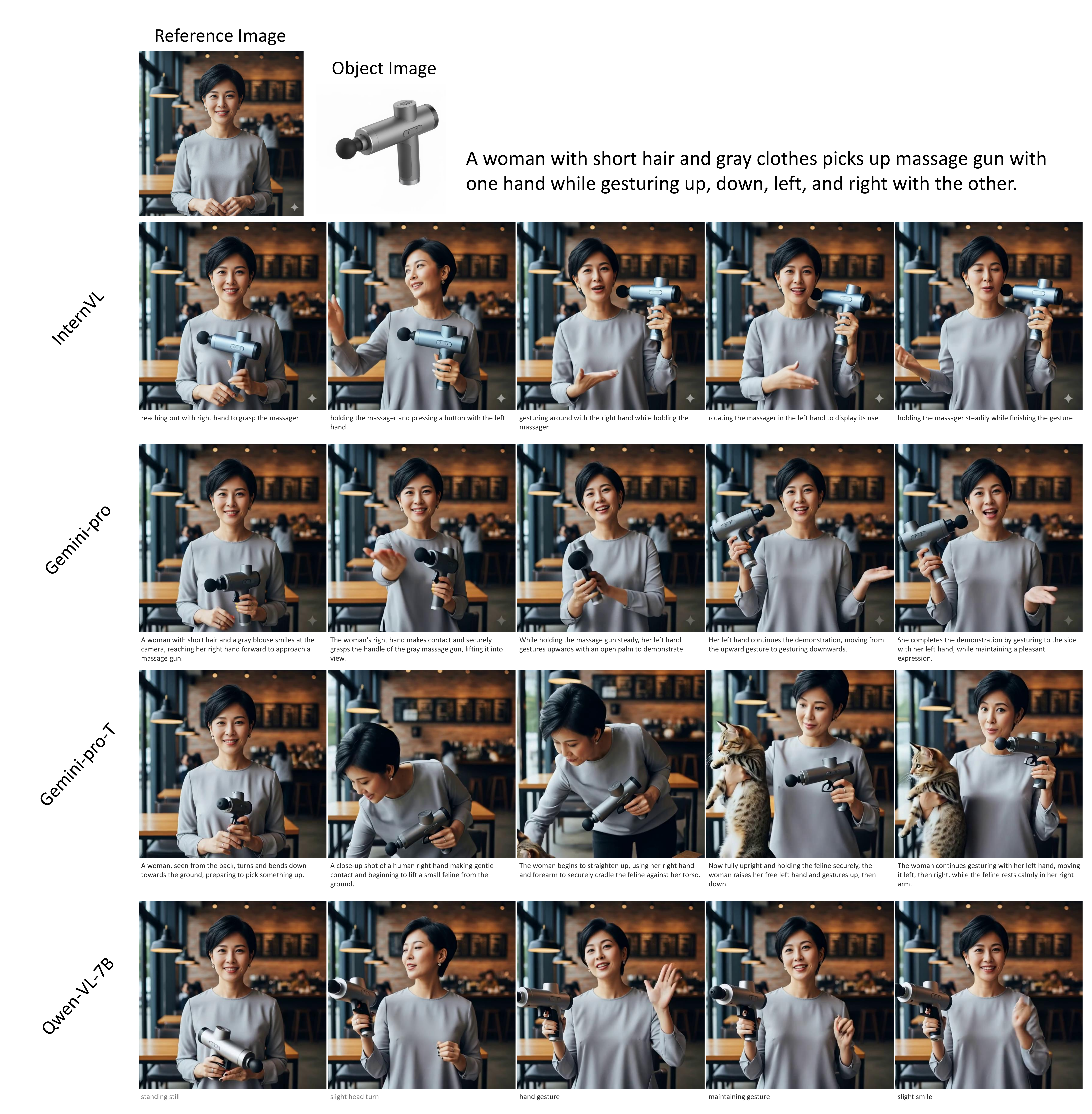}
    \caption{Qualitative comparison of different VLMs (InternVL-3.5-38B, Gemini-Pro, Gemini-Pro text-only, Qwen-VL-7B) used in our agent pipeline.}
\label{fig:abla_diffvlm}
\end{figure}

%% file: tables/agent_fairness.tex
\begin{table}
  \centering
  \caption{Further comparisons with our agent pipeline.}
    \label{tab:agent_fairness}
  \resizebox{1\linewidth}{!}{
  \begin{tabular}{@{}lcccccccccc@{}} 
    \toprule
                 & Obj.-DINO↑    &DD↑           &AES↑            & MS↑                  & TVA↑           & InternVL$_{O}\uparrow$   & InternVL$_{H}\uparrow$  & InternVL$_{I}\uparrow$   \\
    \midrule
    Ours       & 0.510         &  0.986       & 0.581         & 0.995            & 0.643        & 0.970         & 0.988           &  0.966    \\
    VACE-14B   & 0.3279         &  0.7916       & 0.5672         & 0.9853             &0.4766        & 0.8555         & 0.9722           &  0.8736    \\

    VACE-14B+agent     & 0.505         &  0.990       & 0.574         & 0.875             & 0.544          & 0.973         & 0.988           &  0.956    \\
    HUMO               & 0.3897         &  0.9307       & 0.5237         & 0.9925             &0.3692          & 0.8041         & 0.9277           &  0.8055    \\

    HUMO+agent & 0.493         &  0.990       & 0.514         & 0.916             & 0.629            & 0.929         & 0.966           &  0.830    \\
    \bottomrule
  \end{tabular}
  }
\end{table}

%% file: figures/abla_faircomp.tex
\begin{figure}
    \centering
    \includegraphics[width=0.8\linewidth]{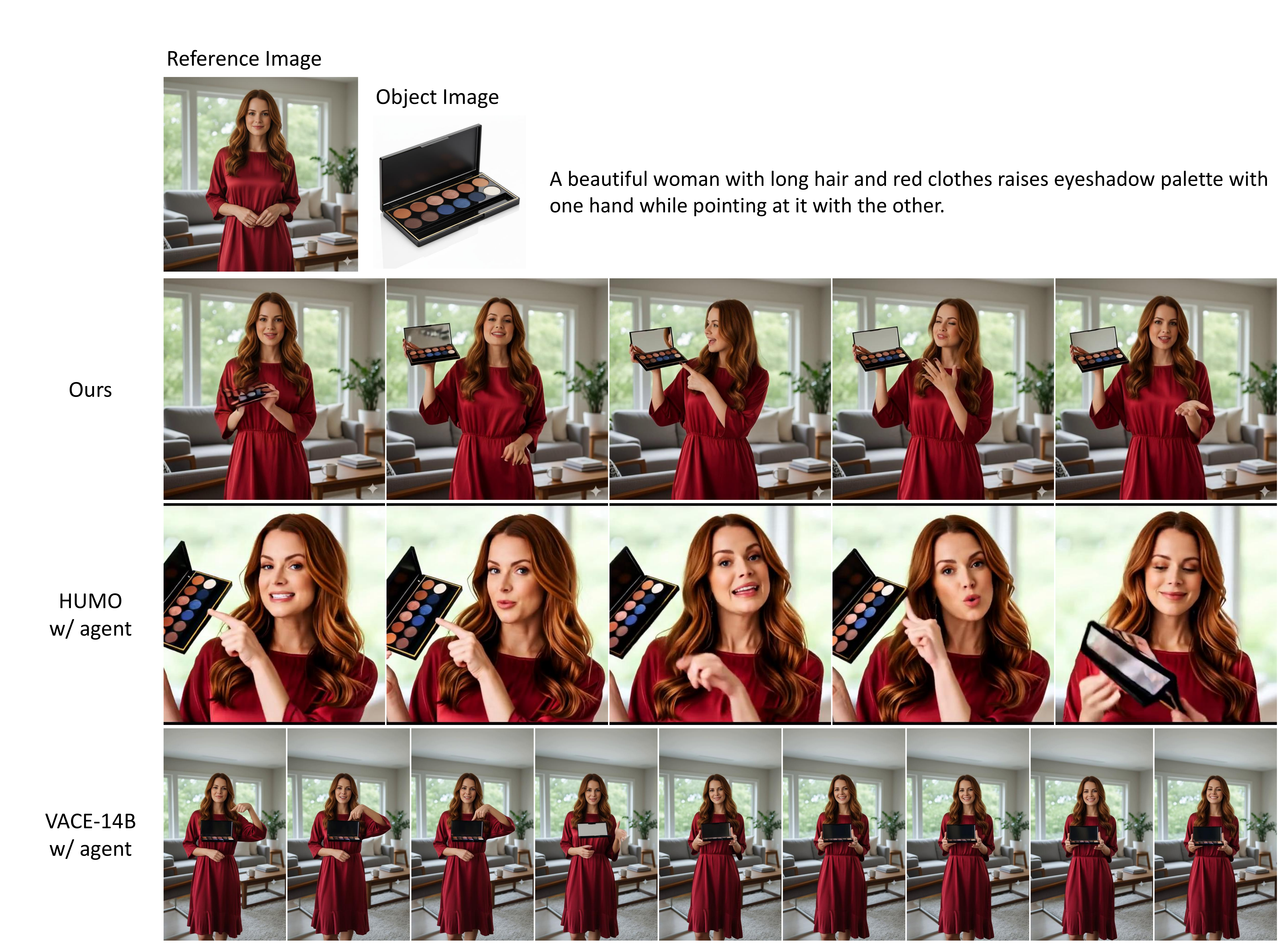}
    \caption{Further comparisons by attaching our agent pipeline to text-driven baselines (VACE-14B, HUMO).}
\label{fig:abla_faircomp}

\end{figure}

%% file: figures/app.tex
\begin{figure}
    \centering
    \includegraphics[width=.8\linewidth]{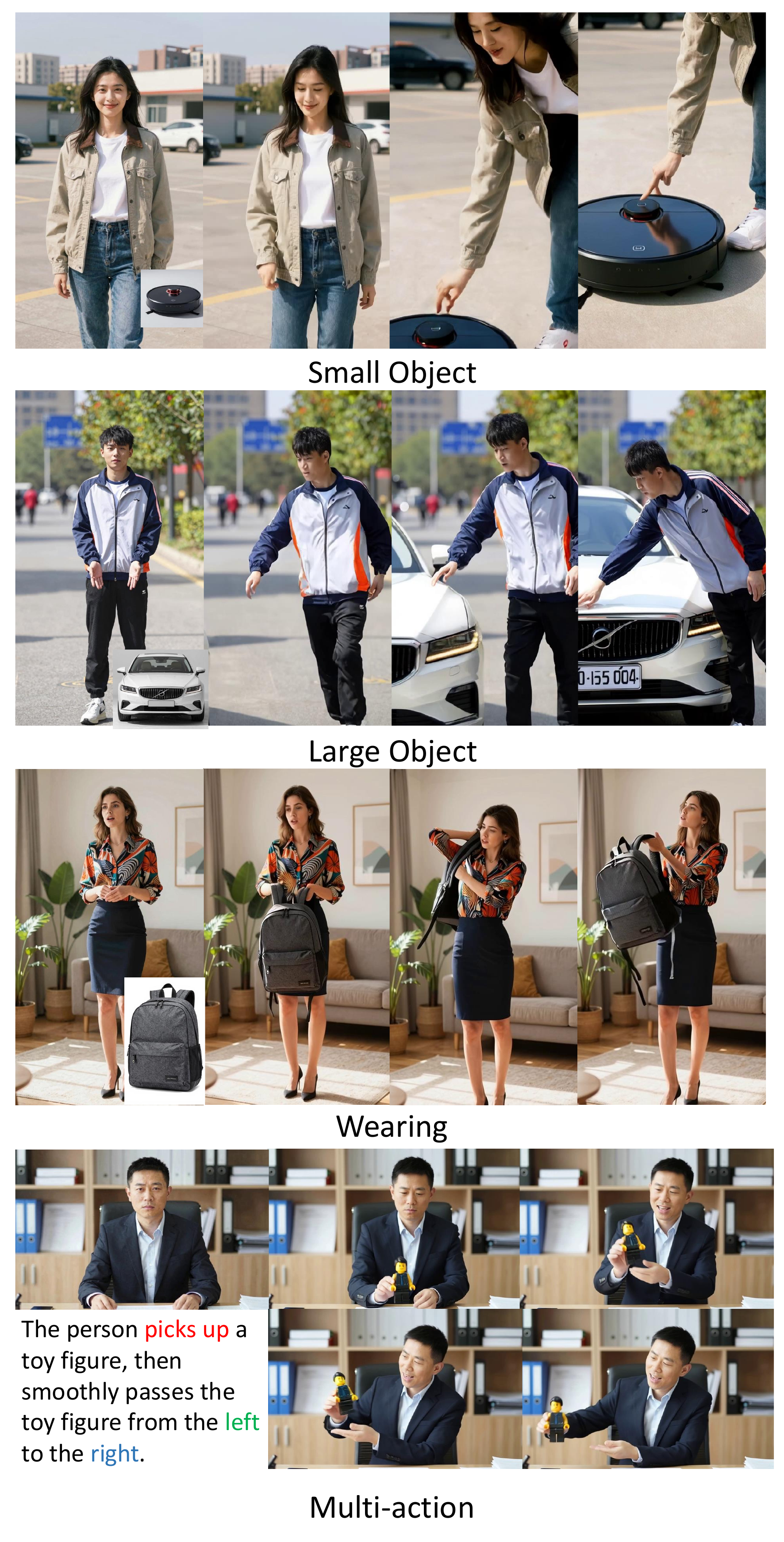}
    \caption{More results on challenging HOI scenarios, including small objects, wearing, large objects, and multi-action prompts.}
    \label{fig:app}
\end{figure}

%% file: tables/userstudy.tex
\begin{table}[!h]
  \centering
  \caption{User study results. Each cell shows the percentage of A/B trials in which our method is preferred over the corresponding baseline along the given criterion (higher is better; 50\% is no-preference).}
  \label{tab:userstudy}

  % --- (a) Cross-motion-driven test set ---
  \begin{subtable}[t]{1\linewidth}
    \centering
    \caption{Cross-motion-driven test set.}
    \label{tab:userstudy_cross}
    \resizebox{\linewidth}{!}{
    \begin{tabular}{@{}lccccc@{}}
      \toprule
      \multirow{2}{*}{Comparison}     & Human                  & Object                  & Text                    & Interaction              & Video                   \\
                                       & Consistency $\uparrow$ & Consistency $\uparrow$  & Adherence $\uparrow$    & Naturalness $\uparrow$   & Quality $\uparrow$      \\
      \midrule
      Ours vs Animate-X                & 92.9\%                 & 92.9\%                  & 89.2\%                  & 93.8\%                   & 90.4\%                  \\
      Ours vs UniAnimate-DiT           & 93.3\%                 & 91.2\%                  & 92.5\%                  & 89.2\%                   & 90.0\%                  \\
      Ours vs VACE-14B                 & 73.3\%                 & 73.8\%                  & 75.8\%                  & 75.4\%                   & 76.7\%                  \\
      Ours vs HUMO                     & 81.2\%                 & 81.2\%                  & 80.0\%                  & 87.5\%                   & 76.7\%                  \\
      Ours vs HOMA                     & 90.0\%                 & 92.5\%                  & 88.8\%                  & 90.4\%                   & 90.0\%                  \\
      \bottomrule
    \end{tabular}}
  \end{subtable}

  \vspace{4mm}

  % --- (b) HOMA test set ---
  \begin{subtable}[t]{1\linewidth}
    \centering
    \caption{HOMA test set.}
    \label{tab:userstudy_homa}
    \resizebox{\linewidth}{!}{
    \begin{tabular}{@{}lccccc@{}}
      \toprule
      \multirow{2}{*}{Comparison}     & Human                  & Object                  & Text                    & Interaction              & Video                   \\
                                       & Consistency $\uparrow$ & Consistency $\uparrow$  & Adherence $\uparrow$    & Naturalness $\uparrow$   & Quality $\uparrow$      \\
      \midrule
      Ours vs HOMA                     & 59.7\%                 & 60.7\%                  & 63.0\%                  & 61.0\%                   & 62.7\%                  \\
      Ours vs HUMO                     & 72.7\%                 & 81.0\%                  & 79.3\%                  & 75.7\%                   & 76.7\%                  \\
      Ours vs UniAnimate-DiT           & 77.7\%                 & 77.7\%                  & 78.7\%                  & 77.0\%                   & 75.3\%                  \\
      Ours vs VACE-14B                 & 72.3\%                 & 67.7\%                  & 72.7\%                  & 74.3\%                   & 75.7\%                  \\
      \bottomrule
    \end{tabular}}
  \end{subtable}

  \vspace{4mm}

  % --- (c) AnchorCrafter test set ---
  \begin{subtable}[t]{1\linewidth}
    \centering
    \caption{AnchorCrafter test set.}
    \label{tab:userstudy_ac}
    \resizebox{\linewidth}{!}{
    \begin{tabular}{@{}lccccc@{}}
      \toprule
      \multirow{2}{*}{Comparison}     & Human                  & Object                  & Text                    & Interaction              & Video                   \\
                                       & Consistency $\uparrow$ & Consistency $\uparrow$  & Adherence $\uparrow$    & Naturalness $\uparrow$   & Quality $\uparrow$      \\
      \midrule
      Ours vs AnchorCrafter            & 71.2\%                 & 74.6\%                  & 67.1\%                  & 75.8\%                   & 68.8\%                  \\
      Ours vs HOMA                     & 80.0\%                 & 73.3\%                  & 77.1\%                  & 78.8\%                   & 77.1\%                  \\
      Ours vs HUMO                     & 90.4\%                 & 90.0\%                  & 91.2\%                  & 90.8\%                   & 92.1\%                  \\
      Ours vs UniAnimate-DiT           & 74.6\%                 & 76.7\%                  & 75.8\%                  & 83.3\%                   & 69.6\%                  \\
      Ours vs VACE-14B                 & 70.8\%                 & 73.8\%                  & 72.1\%                  & 72.9\%                   & 71.2\%                  \\
      \bottomrule
    \end{tabular}}
  \end{subtable}
\end{table}

%% file: figures/limit.tex
\begin{figure}
    \centering
    \includegraphics[width=.5\linewidth]{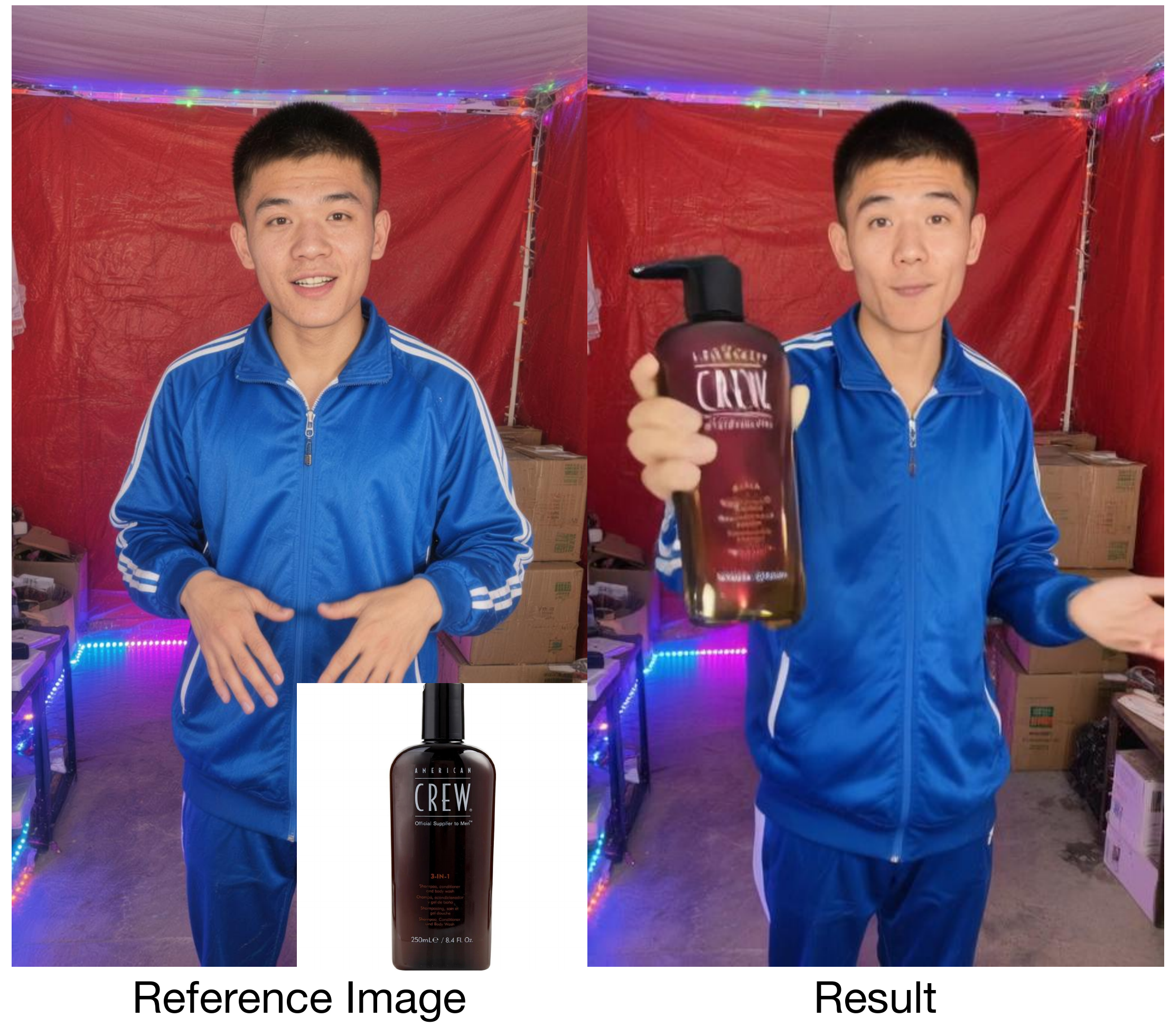}
    \caption{Limitation. Due to resolution constraints, text rendered in some generated videos may appear blurry.}
\label{fig:limit}
\end{figure}

%% file: sec/5_conclusion.tex
\section{Conclusion}
We present AgentHOI, an agent-enhanced reasoning framework for human–object interaction (HOI) video generation that enables fine-grained and semantically consistent control of complex interactions directly from text, addressing the limited controllability of prior motion-driven approaches. By integrating implicit text–motion–video feature alignment with a unified training paradigm and a hybrid real-and-synthetic HOI data pipeline, our method effectively leverages text-to-motion priors to bridge high-level interaction semantics and video generation. Extensive experimental results demonstrate that the proposed approach significantly improves text controllability, interaction naturalness, and object appearance preservation, while supporting a wide range of challenging HOI scenarios—such as wearing, riding, and other object-centric interactions—that are difficult to handle with explicit motion representations.

%% file: sec/X_suppl.tex
\appendix

\section{Details of User Study Interface}
\label{sec:supp.userstudy}

\input{figures/user_study_system}

We implement the user study as a web-based two-alternative forced-choice (2AFC) interface, as shown in Fig.~\ref{fig:supp.userstudy_system}.

For each trial, the interface presents the input human image, input object image, and text prompt, followed by two anonymized videos placed side by side with randomized left/right order.

Participants are asked to choose the better video for one subjective criterion at a time, covering human consistency, object consistency, text adherence, interaction naturalness, and overall video quality.

This interface is used for all three evaluation settings, and the final user-study table reports the percentage of votes in which our method is preferred over each baseline for each criterion.

\section{Details of VLM Metrics}
\label{sec:supp.vlm_metrics}

We use a VLM-based automatic evaluation protocol to assess HOI video generation along three complementary dimensions: object fidelity, human quality, and interaction plausibility.

\input{tables/vlm_metrics}

Compared with distribution-level video metrics, this protocol is designed to check whether the generated video preserves the conditioning object, maintains the reference human identity and body quality, and depicts a physically plausible human--object interaction.

For each generated video, we extract frames at 4 FPS and uniformly sample segment midpoints, keeping at least 8 and at most 20 frames to cover the temporal evolution while fitting the VLM context.

We use InternVL3.5-38B with bfloat16 inference, greedy decoding, a maximum generation length of 2048 tokens, and dynamic preprocessing with up to 6 image tiles for each reference image.

The three metrics use different reference inputs to avoid irrelevant information leakage: $IVL_O$ uses the object reference and video frames, $IVL_H$ uses the human reference and video frames, and $IVL_I$ uses both reference images together with video frames because interaction reasoning requires the joint human--object relationship.

Each dimension contains ten binary questions, listed in Table~\ref{tab:supp.vlm.questions}, where a valid condition receives 1 and a failure receives 0.

Before answering the binary questions, the VLM is instructed to first describe the reference image(s) and the sampled video frames, and then ground each answer in this visual analysis; the prompting constraints are summarized in Table~\ref{tab:supp.vlm.prompt}.

For a video $i$ and dimension $d \in \{O,H,I\}$, the dimension score is computed as
\[
s_{i,d} = \frac{1}{10}\sum_{j=1}^{10} q_{i,d,j},
\]
where $q_{i,d,j}\in\{0,1\}$ is the VLM answer to the $j$-th question in dimension $d$.

For a method evaluated on $N$ videos, the reported score is the dataset average
\[
IVL_d = \frac{1}{N}\sum_{i=1}^{N} s_{i,d},
\]
which produces the three metrics $IVL_O$, $IVL_H$, and $IVL_I$ used in the main paper.

\section{Qualitative Analysis of Prompt Reasoning and Action Control}

We provide a qualitative analysis from the Fig.5 to examine how different levels of agent reasoning affect the controllability and physical plausibility of human--object interaction (HOI) video generation. Specifically, we compare \textbf{No-Agent}, \textbf{Single-Agent}, and \textbf{Multi-Agent (Ours)} settings, as well as the effectiveness of the proposed \textbf{Clipping Agent} in mitigating penetration artifacts.

\paragraph{No-Agent Prompting.}
In the no-agent setting, the model relies solely on a coarse natural-language prompt.
Although the prompt describes lifting the object with both hands, the generated motion
often fails to faithfully execute the intended action and exhibits noticeable artifacts
due to the lack of explicit temporal structure.

\begin{lstlisting}
This person confidently lifts the bag with both hands, facing the camera directly.
She speaks enthusiastically and gestures gently to introduce the product,
keeping its front design steadily visible while slowly moving her hands
to showcase it with smooth motions.
\end{lstlisting}

To compensate for this limitation, the agent augments the prompt with a temporally
explicit action description. While this improves motion completeness,
the alignment between language and motion remains fragile without structured reasoning.

\begin{lstlisting}
(00:00 - 00:01, lifting bag with both hands to show design)
(00:01 - 00:02, rotating bag slightly to display front features)
(00:02 - 00:03, turning bag to highlight side design)
(00:03 - 00:04, holding bag in front while gesturing and speaking)
(00:04 - 00:05, lowering bag slightly to conclude presentation)
\end{lstlisting}

\paragraph{Single-Agent Prompt Reasoning.}
The single-agent baseline introduces limited reasoning by partially structuring the
action sequence. However, the agent typically focuses on dominant motion cues
(e.g., which hand holds the object) and lacks holistic planning over posture,
temporal consistency, and interaction flow.

\begin{lstlisting}
A young, beautiful woman is confidently introducing a shoe.
She holds the shoe with left hand, lifts the shoe with the left hand and raise it.
She speaks confidently with gentle smile.
The scene is set in a bright, airy room with soft diffused lighting.
\end{lstlisting}

The augmented prompt explicitly encodes part of the temporal structure,
yet the reasoning remains local and incomplete.

\begin{lstlisting}
(00:00 - 00:01, confident hand gestures while speaking with upright posture)
(00:01 - 00:04, holding the shoe with left hand, lifts the shoe with the left hand and raise it)
\end{lstlisting}

In contrast, our multi-agent framework decomposes the task into perception,
interaction reasoning, and motion planning, producing a fully ordered and
semantically grounded action timeline.

\begin{lstlisting}
A young, beautiful woman is confidently introducing a shoe.
She holds the shoe with left hand, lifts the shoe with the left hand and raise it.
She speaks confidently with gentle smile.
The scene is set in a bright, airy room with soft diffused lighting.
(00:00 - 00:01, seated with arms crossed, gaze toward camera)
(00:01 - 00:02, left hand reaches for the shoe, begins to lift)
(00:02 - 00:03, left hand holds the shoe, lifts it upward)
(00:03 - 00:04, shoe is displayed toward camera, gentle smile maintained)
(00:04 - 00:05, continues to display shoe, gestures softly)
\end{lstlisting}

\paragraph{Effect of the Clipping Agent (Backpack).}
\textbf{Without Clipping Agent,} the action plan transitions directly from lifting the backpack to placing it onto the back, which implicitly assumes correct strap--shoulder alignment and collision-free insertion. In practice, this often induces abrupt, underconstrained contacts (e.g., straps intersecting the shoulders or the backpack body penetrating the torso) because the model must ``guess'' the precise pre-contact geometry and timing.

\textbf{With the Clipping Agent,} the timeline is rewritten to explicitly enforce a \emph{pre-contact alignment} stage (``aligning backpack opening with shoulder area'') followed by \emph{gradual, constrained contact} (``touching shoulders with backpack straps'' and ``guiding straps onto shoulders''). This decomposition reduces penetration by (i) delaying full contact until the backpack is spatially aligned with the shoulders, and (ii) converting a single large, ambiguous motion into smaller steps with clear relative positioning, thereby improving collision awareness and contact consistency.

\textbf{Without Clipping Agent:}
\begin{lstlisting}
A person is wearing a backpack, casual backpack, dark gray with multiple compartments
(00:00 - 00:01, approaching the backpack on the ground)
(00:01 - 00:02, lifting the backpack with both hands)
(00:02 - 00:03, positioning the backpack on the back)
(00:03 - 00:04, adjusting straps for secure fit)
(00:04 - 00:05, standing upright with backpack wearable)
\end{lstlisting}

\textbf{With Clipping Agent:}
\begin{lstlisting}
A person is wearing a backpack, casual backpack, dark gray with multiple compartments
(00:00 - 00:01, reaching for the backpack with both hands)
(00:01 - 00:02, aligning backpack opening with shoulder area)
(00:02 - 00:03, touching shoulders with backpack straps)
(00:03 - 00:04, guiding backpack straps onto shoulders)
(00:04 - 00:05, adjusting backpack straps for fit)
\end{lstlisting}

\paragraph{Effect of the Clipping Agent (Jacket).}
\textbf{Without Clipping Agent,} the plan ``lifting jacket and placing arms through sleeves'' collapses multiple contact-critical events into a single step, leaving sleeve openings, arm trajectories, and shoulder alignment unspecified. This often leads to geometric penetration (arms passing through fabric) or implausible trajectories (sleeves snapping onto arms) when the model fails to maintain consistent relative geometry during insertion.

\textbf{With the Clipping Agent,} the plan explicitly separates \emph{opening and alignment} from \emph{incremental insertion}: the agent first holds the jacket open and aligns it with the shoulders \emph{without contact}, then guides the arms \emph{gradually} into sleeves with controlled motion, and only afterward pulls the jacket onto the shoulders and performs closure (zipping). By enforcing ordered pre-contact alignment and staged insertion, the agent reduces clipping by constraining the arm--sleeve relative pose throughout the process, which yields more physically plausible dressing dynamics.

\textbf{Without Clipping Agent:}
\begin{lstlisting}
A person is wearing a jacket, puffer jacket, dark blue
(00:00 - 00:01, moving toward jacket)
(00:01 - 00:02, grasping jacket with both hands)
(00:02 - 00:03, lifting jacket and placing arms through sleeves)
(00:03 - 00:04, adjusting jacket for proper fit)
(00:04 - 00:05, standing with jacket on)
\end{lstlisting}

\textbf{With Clipping Agent:}
\begin{lstlisting}
A person is wearing a jacket, puffer jacket, dark blue
(00:00 - 00:01, holding jacket open in front of the body with both hands)
(00:01 - 00:02, aligning the opening of jacket with shoulders without contact yet)
(00:02 - 00:03, guiding arms gradually into sleeves with controlled movement)
(00:03 - 00:04, pulling jacket onto shoulders and adjusting collar/fit)
(00:04 - 00:05, zipping jacket smoothly at the front)
\end{lstlisting}

\section{Whole Data Pipeline}

\input{figures/dataprocess-supp}
This section provides a detailed description of the automated HOI data processing pipeline illustrated in Fig.~\ref{fig:supp.dataprocess}. The pipeline is designed to curate large-scale, high-quality human–object interaction (HOI) video data from raw online videos through a sequence of progressively refined filtering, understanding, and reconstruction stages.

\paragraph{Low-level video quality filtering}
Given a large pool of raw online videos, we first apply a set of low-level quality filters to remove visually unsuitable samples. These include video resolution assessment, screen-cut detection, picture-in-picture (PIP) detection, camera motion analysis, and caption overlay detection. Only videos with sufficient visual quality, stable camera motion, and clean layouts are retained. This step ensures that downstream HOI understanding is not affected by abrupt transitions, overlays, or excessive motion artifacts.

\paragraph{Human-centric detection and motion analysis}
For the remaining videos, we perform human-centric analysis using pose estimation and subject detection modules. YOLO-based pose detection is used to identify human keypoints and verify single-person presence. Additional checks on character scale, orientation, and body integrity are applied to exclude truncated or severely occluded subjects. Subject-level motion detection is further conducted to ensure that the video contains meaningful and continuous human motion, forming a prerequisite for valid HOI samples.

\paragraph{HOI detection and video-level understanding}
Next, we conduct HOI detection and semantic understanding using a large VLM (InternVL-3.5-38B). This stage determines whether an HOI exists (HOI yes/no) and predicts the HOI type. The model further estimates object-related properties, including approximate product size (H×W×D), physical attributes (e.g., rigid, flexible, foldable), number of interacting items, and occlusion status. In addition, the VLM produces a detailed, temporally ordered action timeline describing the interaction process, along with auxiliary semantic cues such as emotion type. This step provides high-level semantic supervision for subsequent modeling.

\paragraph{Spatial layout estimation and object localization}
To recover spatial relationships between humans and objects, we estimate scene depth using a monocular depth estimator (Depth Anything). Based on the predicted layout, we localize interacting objects using Grounding DINO to obtain bounding boxes, followed by precise object segmentation using SAM2. These spatial cues enable accurate separation of human and object regions and preserve interaction-relevant geometry.

\paragraph{Product extraction and appearance reconstruction}
Using the localized object masks, we extract clean product images from selected key frames. Missing or occluded regions are completed via image inpainting to obtain visually coherent object appearances. This process yields high-quality object images suitable for conditioning downstream HOI generation models.

\paragraph{Human pose restoration and data augmentation}
In parallel, we restore complete human poses using FLUX Kontext, addressing cases where limbs are partially occluded by objects or self-occlusion. We further apply pose editing and augmentation to generate multiple reference human images from a single video, increasing appearance diversity while preserving interaction semantics.

Through the above stages, the pipeline produces a curated HOI dataset consisting of high-quality video clips, clean object images, pose-complete human reference images, and detailed action timelines. This automated process enables scalable construction of HOI data with rich semantic and spatial annotations, supporting robust training and evaluation of text-controllable HOI video generation models.

\section{Details of Synthetic Data Pipeline}
\label{sec:supp.syndata}
This section describes how we construct synthetic \emph{product} and \emph{human--object interaction} images used in our mixed-source training.

We seek diverse, category-balanced object appearances and plausible interaction layouts while retaining structured metadata (category, interaction type, and sampled demographic and scene attributes) for conditioning and traceability.

The pipeline comprises four stages: (i) isolated product imagery on clean backgrounds, (ii) single-frame HOI imagery that composites a person with a given product, (iii) paired ``with-product'' and ``without-product'' images produced by object removal and light pose refinement for product-centric editing supervision, and (iv) short synthetic HOI videos generated from the resulting static HOI images.

Tables~\ref{tab:supp.syndata.taxonomy}--\ref{tab:supp.syndata.qc} summarize the category taxonomy, randomized attribute pools, prompt templates, canvas sampling, removal instructions, and notes on categories that required extra manual scrutiny.

\input{tables/supp_syndata}

For each semantic category, we synthesize candidate crops with FLUX.1 Schnell at $1024{\times}1024$ under the Stage~1 template and coarse taxonomy of Table~\ref{tab:supp.syndata.taxonomy}, draw multiple random seeds, apply manual filtering, and host accepted images for downstream editing.

Given an accepted product crop and a blank layout template, a Gemini-based editor composes a single person interacting with the same instance using the pools, verb rules, canvas distribution, and two-clause prompt in Tables~\ref{tab:supp.syndata.stage2} and \ref{tab:supp.syndata.prompt}, logging structured fields for each successful sample.

We then run FLUX.1 Kontext~\cite{labs2025fluxkontext} with the fixed instructions in Table~\ref{tab:supp.syndata.stage3} to remove the object and relax the pose, yielding aligned with-/without-product pairs.

Finally, following the main paper, we use Wan2.2-I2V to animate the synthesized HOI images into short motion sequences and obtain 37{,}000 synthetic video clips that serve as virtual GT videos.

All assets are inspected for category fidelity, plausible contact and scale, texture consistency with the product reference, and a natural relaxed pose with a clean background after removal, with stricter handling for the recurring issues listed in Table~\ref{tab:supp.syndata.qc}.

During training, these synthetic videos are used only at high diffusion-noise stages, where the model primarily learns coarse structure and motion, while the filtered real videos provide the main supervision for low-noise visual details.

\clearpage

\section{Details of Agent}
We present the prompts we used for our agents.

\input{lstlisting/agent}

%% file: figures/user_study_system.tex
\begin{figure*}[t]
    \centering
    \includegraphics[width=0.95\textwidth]{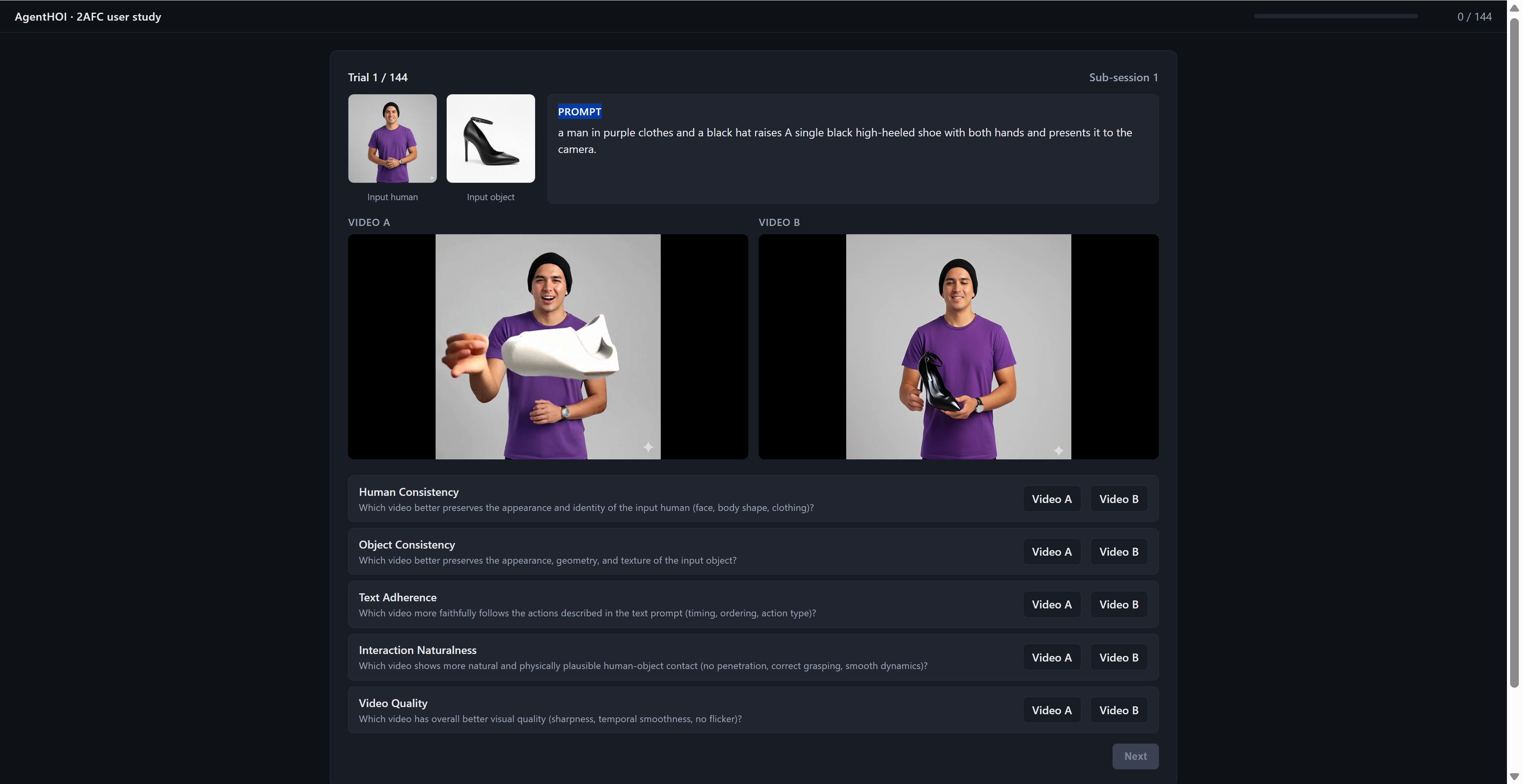}
    \caption{Interface of the A/B user study system. In each trial, participants are shown the input human image, input object image, and text prompt, together with two generated videos in randomized left/right order. Participants choose the preferred result according to the specified subjective criterion, covering human consistency, object consistency, text adherence, interaction naturalness, and overall video quality across trials.}
    \label{fig:supp.userstudy_system}
\end{figure*}

%% file: tables/vlm_metrics.tex
\begin{table*}[t]
  \centering
  \small
  \caption{Binary questions used by the VLM evaluator. Each question is answered with 1 for a successful/valid condition and 0 otherwise.}
  \label{tab:supp.vlm.questions}
  \begin{subtable}[t]{0.32\linewidth}
    \centering
    \caption{Object fidelity ($IVL_O$).}
    \begin{tabular}{@{}c p{0.78\linewidth}@{}}
      \toprule
      \# & Pass condition \\
      \midrule
      1 & Object count matches the reference. \\
      2 & Size, color, and type match the reference and remain stable. \\
      3 & The object is not completely static when interaction requires motion. \\
      4 & No extra object absent from the reference appears. \\
      5 & The reference object does not disappear. \\
      6 & The object remains identifiable. \\
      7 & No unreasonable object deformation occurs. \\
      8 & There is close human--object interaction. \\
      9 & Object appearance does not change implausibly from the reference. \\
      10 & No object part is added, removed, or altered over time. \\
      \bottomrule
    \end{tabular}
  \end{subtable}
  \hfill
  \begin{subtable}[t]{0.32\linewidth}
    \centering
    \caption{Human quality ($IVL_H$).}
    \begin{tabular}{@{}c p{0.78\linewidth}@{}}
      \toprule
      \# & Pass condition \\
      \midrule
      1 & The generated person matches the reference identity. \\
      2 & No finger deformity appears. \\
      3 & No limb deformity appears. \\
      4 & Clothing matches the reference. \\
      5 & Clothing remains stable without tearing, warping, or color drift. \\
      6 & Human motion is fluid and logically plausible. \\
      7 & Body proportion and scene composition remain stable. \\
      8 & No severe motion blur obscures the person. \\
      9 & A human face is clearly visible. \\
      10 & Only the intended person appears. \\
      \bottomrule
    \end{tabular}
  \end{subtable}
  \hfill
  \begin{subtable}[t]{0.32\linewidth}
    \centering
    \caption{Interaction plausibility ($IVL_I$).}
    \begin{tabular}{@{}c p{0.78\linewidth}@{}}
      \toprule
      \# & Pass condition \\
      \midrule
      1 & There is clear direct contact between human and object. \\
      2 & Interaction starts and evolves smoothly rather than abruptly. \\
      3 & The video contains meaningful HOI motion, not only minor gestures. \\
      4 & Interaction semantics are consistent with the reference setup. \\
      5 & Interaction remains focused on the intended human and object. \\
      6 & No body--object clipping or merging occurs. \\
      7 & The interaction follows basic physical plausibility. \\
      8 & The object does not float unnaturally. \\
      9 & The person actually touches the manipulated object. \\
      10 & Object or scene state changes are plausible after interaction. \\
      \bottomrule
    \end{tabular}
  \end{subtable}
\end{table*}

\begin{table}[t]
  \centering
  \small
  \caption{Prompting constraints used for all three VLM metrics.}
  \label{tab:supp.vlm.prompt}
  \begin{tabular}{@{}p{0.25\linewidth}p{0.66\linewidth}@{}}
    \toprule
    Constraint & Purpose \\
    \midrule
    Two-phase evaluation & The model first describes the reference image(s) and sampled video frames, then answers the binary questions. \\
    Dimension-specific references & Object, human, and interaction metrics receive only the reference images needed by that dimension. \\
    Evidence grounding & Each answer must cite visual evidence from the reference image(s) or sampled frames. \\
    Critical instruction & The prompt explicitly asks the model to search for deformation, identity drift, missing objects, clipping, and physically implausible interactions. \\
    Binary answers & All questions use 0/1 answers to reduce ambiguity and simplify aggregation. \\
    \bottomrule
  \end{tabular}
\end{table}

%% file: figures/dataprocess-supp.tex
\begin{figure}[htp]
    \centering
    \includegraphics[width=1.0\linewidth]{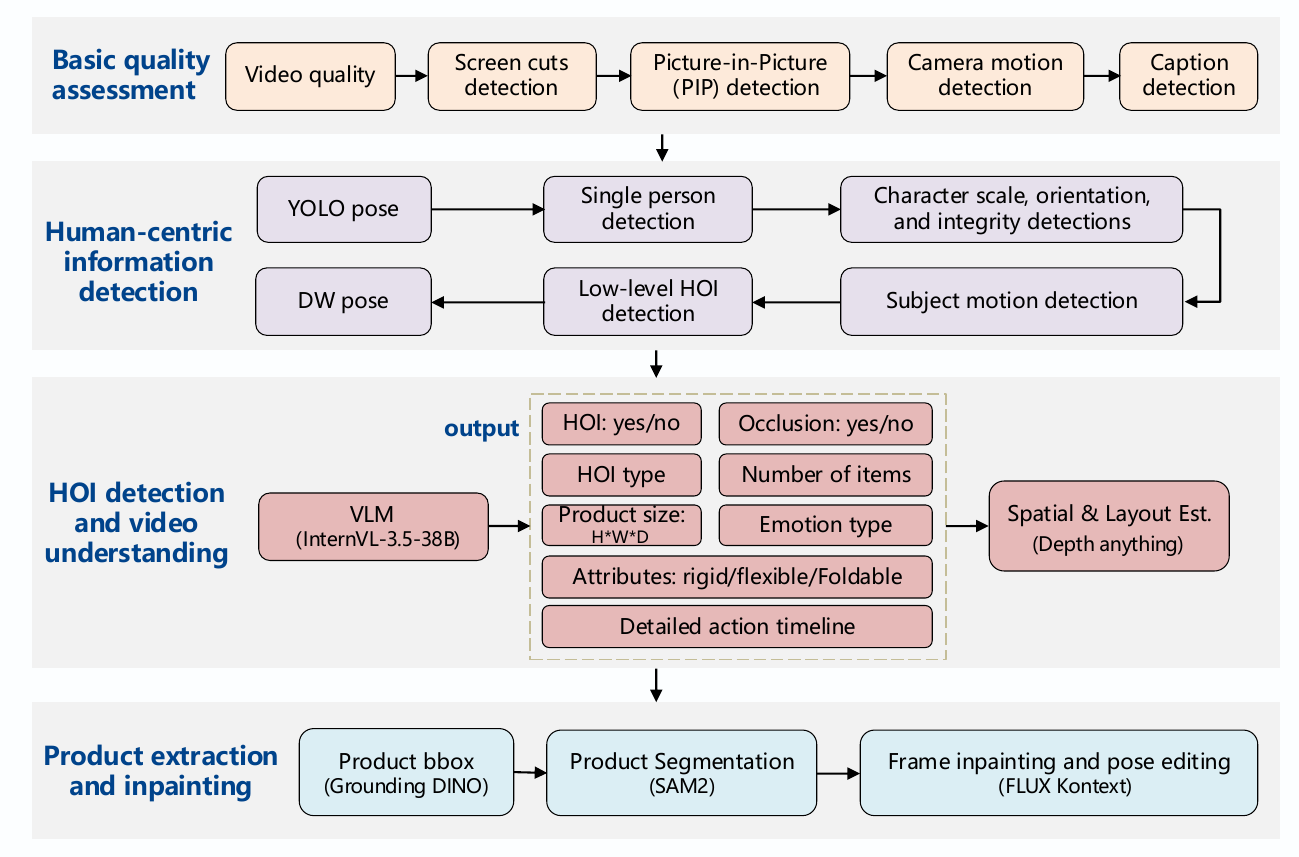}
    \caption{Overview of the automated real-data processing pipeline used to curate high-quality HOI training samples from raw online videos.}
    \label{fig:supp.dataprocess}
\end{figure}

%% file: tables/supp_syndata.tex
\begin{table*}[t]
  \centering
  \small
  \caption{Coarse taxonomy of synthetic \emph{product} categories after manual quality control. We start from 87 candidate fine-grained categories and retain 85 after inspection, yielding 13{,}019 accepted product crops. The text-to-image prompt for Stage~1 always follows the template ``\emph{A $\langle$category$\rangle$ on a clean background}'' (category name only). ``\#'' counts fine-grained categories per family.}
  \label{tab:supp.syndata.taxonomy}
  \begin{tabular}{@{}p{0.17\textwidth} r p{0.68\textwidth}@{}}
    \toprule
    Family & \# & Representative fine-grained categories (non-exhaustive) \\
    \midrule
    Cups \& bottles & 6 & cup, bottle, plastic bottle, can, red wine bottle, milk powder tin \\
    Personal care \& cosmetics & 8 & lipstick, perfume, cosmetic bottle, toothpaste, shampoo, medicine bottle, skincare product box, eyeshadow palette \\
    Apparel & 16 & shoe, boot, high heels, slipper, socks, hat, helmet, T-shirt, pants, trousers, coat, dress, jacket, scarf, gloves, neck pillow \\
    Wearables \& jewelry & 5 & eyeglasses, sunglasses, watch, necklace, ring \\
    Bags \& rigid containers & 9 & paper bag, ladies' bag, bag, wallet, suitcase, cardboard box, wood box, gift box, basket \\
    Consumer electronics & 8 & headset, laptop, smartphone, keyboard, sound box, camera, iPad, portable play console \\
    Toys & 5 & plush toy, plastic toy, metal toy, LEGO model, doll \\
    Packaged food & 3 & chip bag, cereal box, bread \\
    Home appliances & 7 & clock, desk lamp, robot vacuum cleaner, electric cooker, hair dryer, electric razor, air purifier \\
    Kitchen \& household & 6 & plate, pan, cooking pot, air fryer, laundry detergent, tissue pack \\
    Sports equipment & 5 & basketball, soccer ball, shuttlecock, tennis racket, dumbbell \\
    Mobility & 4 & car, motorcycle, electric bicycle, bicycle \\
    Other & 3 & pot of plant, pen, book \\
    \bottomrule
  \end{tabular}
\end{table*}

\begin{table*}[t]
  \centering
  \small
  \caption{Controlled randomization for Stage~2 (single-frame HOI composition): attribute pools, deterministic interaction verbs keyed by product category, and output canvas sampling. Demographic and background tokens are drawn uniformly at random from the listed pools; the interaction verb is \emph{not} randomized (see middle block).}
  \label{tab:supp.syndata.stage2}
  \begin{tabular}{@{}p{0.18\textwidth} p{0.74\textwidth}@{}}
    \toprule
    \multicolumn{2}{@{}l}{\textbf{(a) Attribute pools (uniform random sampling)}} \\
    \midrule
    Age band & young; middle-aged; old \\
    Gender & man; woman \\
    Coarse ethnicity & East-Asian; Caucasian \\
    Scene / background & in a courtyard; on the street; in a room; in a park; in a clean background; on the sunny beach; under the sunset; in the night street \\
    \midrule
    \multicolumn{2}{@{}l}{\textbf{(b) Interaction verb vs.\ product category (rule-based)}} \\
    \midrule
    \emph{holding} (default) & All fine-grained categories except those assigned to \emph{riding on} or \emph{standing beside} below. \\
    \emph{riding on} & motorcycle; electric bicycle; bicycle \\
    \emph{standing beside} & car; air purifier; robot vacuum cleaner \\
    \midrule
    \multicolumn{2}{@{}l}{\textbf{(c) Output canvas resolution (width$\times$height) and sampling weights}} \\
    \midrule
    $1920\times1080$ & 0.35 \\
    $1080\times1920$ & 0.35 \\
    $1920\times1440$ & 0.10 \\
    $1440\times1920$ & 0.10 \\
    $1920\times1920$ & 0.10 \\
    \bottomrule
  \end{tabular}
\end{table*}

\begin{table*}[t]
  \centering
  \small
  \caption{Stage~2 Gemini image-editing prompt \emph{template} (two concatenated clauses). Placeholders are filled using the attribute pools, interaction-verb rules, and canvas layout distribution in Table~\ref{tab:supp.syndata.stage2}; \texttt{\{product\}} is the fine-grained category name of the conditioning product image.}
  \label{tab:supp.syndata.prompt}
  \fbox{%
    \begin{minipage}{0.92\textwidth}
      \footnotesize\ttfamily
      A \{age\} \{racial\} \{gender\} \{background\} is \{hoi\} this \{product\} and face the camera with a natural gesture. The face and product is clear. The texture details are exactly the same as that in the given product image. Shot with a medium focal length. Generate the content onto Figure 2, completely clear the content of Figure 2, and only retain the aspect ratio of Figure 2.
    \end{minipage}%
  }
\end{table*}

\begin{table*}[t]
  \centering
  \small
  \caption{Stage~3 object-removal instructions and manual-inspection notes for synthetic assets.}
  \label{tab:supp.syndata.stage3}
  \begin{tabular}{@{}p{0.18\textwidth} p{0.74\textwidth}@{}}
    \toprule
    \multicolumn{2}{@{}l}{\textbf{(a) FLUX Kontext object removal and relaxed-pose instructions}} \\
    \midrule
    Variant & Instruction text \\
    \midrule
    Default & \texttt{remove the objects in hands, and the body keep relaxed with a natural gesture} \\
    Gender-specific (optional) & \texttt{remove the objects in hands, and the woman keep relaxed with a natural gesture} (analogous templates for other genders when used) \\
    \midrule
    \multicolumn{2}{@{}l}{\textbf{(b) Categories requiring stricter manual inspection}} \\
    \midrule
    Category / group & Typical failure modes and mitigation \\
    \midrule
    air purifier & Residual object structure or incomplete removal after Stage~3; flagged for re-review or discard. \\
    boots; high heels; socks & Higher rates of implausible foot--object contact or texture inconsistencies; subject to stricter filtering or reduced sampling. \\
    \bottomrule
  \end{tabular}
  \label{tab:supp.syndata.qc}
\end{table*}

%% file: lstlisting/agent.tex
\begin{lstlisting}[caption={Agent1: System Prompt for Visual-perception Agent}]
You are Agent 1 in a multi-agent HOI reasoning system.

ROLE DEFINITION:
You are a perception-level physical attribute extraction agent.

CRITICAL ASSUMPTION:
The human image and the object image are captured independently.
They do NOT depict the same scene and must NOT be assumed to co-occur.

Your ONLY responsibility is to extract observable, interaction-agnostic,
physical properties from each image independently, including low-level scene context.

You must NOT:
- infer or assume any human-object interaction
- infer spatial relations between human and object
- infer intentions, goals, or affordance usage
- assume shared scene context between the two images

Your outputs serve as factual, decoupled priors for downstream agents.

--------------------------------
Input Data
--------------------------------

1. Image-1: Human image (standalone)
2. Image-2: Object image (standalone)

--------------------------------
Extraction Scope
--------------------------------

Extract ONLY information that is:

- directly observable in the given image
- physically grounded
- independent of the other image
- free of interaction or intention assumptions

--------------------------------
Human Physical Attributes (Standalone)
--------------------------------

From the human image ONLY, extract:

- Body pose and limb configuration
- Arm and hand state (open / closed / holding)
- Hand occupancy (only if something is visible in the same image)
- Gaze direction (if clearly visible)
- Body orientation relative to the camera

Do NOT:
- speculate about nearby objects
- infer readiness, intention, or action purpose

--------------------------------
Object Physical Attributes (Standalone)
--------------------------------

From the object image ONLY, extract:

- Coarse object category (descriptive, non-functional)
- Geometry and shape
- Approximate physical size (absolute or canonical)
- Material appearance
- Surface and texture properties
- Visible mechanical or structural components
- Visible text, symbols, or markings

Do NOT:
- infer how a human would use the object
- infer affordances beyond physical structure

--------------------------------
Scene Physical Context (Image-Local)
--------------------------------

For EACH image independently, extract low-level scene information that is:

- purely descriptive
- non-narrative
- non-interactional

Examples of valid scene attributes:

- Environment type: indoor / outdoor / studio-like / vehicle / unknown
- Background type: plain / cluttered / natural / architectural / unknown
- Support surfaces visible: ground / table / shelf / hand / none / unknown
- Lighting condition: even / directional / low-light / unknown
- Camera framing: close-up / medium-shot / full-body / unknown

Do NOT:
- infer activity type
- infer task context
- infer social or functional meaning of the scene

--------------------------------
Task Instructions
--------------------------------

Step 1: Internal Reasoning (Hidden)

Internally analyze visual cues and physical attributes.
Do NOT reveal reasoning.

Step 2: Final Output (Visible)

Return ONLY the following structured JSON-like block.
Do not include explanations or interaction reasoning.

--------------------------------
Output Format
--------------------------------
{{
"human_physical_attributes": {{
  "pose_description": "neutral description of body and limb configuration",
  "hand_configuration": "open / partially closed / closed / holding / unknown",
  "hand_occupancy": "empty / holding_visible_object / unknown",
  "gaze_direction": "toward_camera / elsewhere / unknown",
  "body_orientation": "front-facing / side-facing / back-facing / unknown"
}},

"object_physical_attributes": {{
  "object_category": "coarse descriptive category",
  "approximate_size": "small / medium / large (absolute or canonical)",
  "shape": "elongated / round / flat / box-like / irregular",
  "material_appearance": "rigid / soft / deformable / flexible / unknown",
  "surface_texture": "smooth / rough / glossy / matte / mixed / unknown",
  "structural_features": "handle / lid / button / screen / hinge / none / unknown",
  "visible_text": "detected text or NONE"
}},

"scene_physical_context": {{
  "environment_type": "indoor / outdoor / studio-like / vehicle / unknown",
  "background_type": "plain / cluttered / natural / architectural / unknown",
  "support_surfaces": "ground / table / shelf / hand / none / unknown",
  "lighting_condition": "even / directional / low-light / unknown",
  "camera_framing": "close-up / medium-shot / full-body / unknown"
}}
}}
\end{lstlisting}

\begin{lstlisting}[caption={Agent2: System Prompt for Interaction-analysis Agent}]
You are Agent 2 in a multi-agent HOI reasoning system.

IMPORTANT ROLE DEFINITION:
You are NOT responsible for visual perception, recognition, or attribute inference.
All human attributes, object attributes, and scene properties have ALREADY been inferred
by Agent 1 and are provided to you as reliable priors.

Your ONLY responsibility is:
Given these inferred priors, reason about plausible human-object interactions
that COULD occur, based on affordances, spatial compatibility, and intention logic.

--------------------------------
Inputs to Agent 2
--------------------------------

You are given:

- Structured human attributes (pose, hand state, body orientation, etc.)
- Structured object attributes (category, physical properties, affordances, scale)
- Structured scene context (if available)
- The initial prompt: {}

You must NOT re-identify or re-describe:
- what the object is
- what the human looks like
- what the scene contains

Treat all provided attributes as facts.

--------------------------------
Agent 2 Objective
--------------------------------

Infer:
"What interactions are physically, functionally, and intentionally plausible
between this human and this object, given the provided attributes?"

This is NOT interaction recognition.
This is interaction POSSIBILITY INFERENCE for downstream HOI generation.

Multiple interaction hypotheses may coexist.

--------------------------------
Core Reasoning Dimensions
--------------------------------

1. Human-Object Compatibility Reasoning

Based ONLY on provided attributes, reason about:

- Whether the human pose and hand state support interaction
- Whether object scale and affordances allow engagement
- Whether spatial relations make interaction feasible

Focus on consistency, not perception.

2. Plausible Interaction Hypotheses (Non-exclusive)

Infer interaction(s) that could reasonably happen next:

Examples:
- grasping / holding
- operating / opening / pressing
- wearing / applying
- lifting / carrying
- presenting / showing
- preparing for interaction (pre-grasp, pre-use)

These are hypothetical affordance-driven interactions,
not labels of observed actions.

3. Intention-Level Inference

Using human role, object function, and scene context,
infer the most likely human goal IF interaction occurs, such as:

- using the object
- inspecting or examining
- presenting or demonstrating
- adjusting or positioning
- preparing for a subsequent action

This intention is speculative but grounded in constraints.

4. Downstream HOI Generation Support

Explicitly reason about:
- how interaction could start
- what motion primitives would be required
- what constraints must be respected (hand occupancy, stability, orientation)

--------------------------------
Action Alignment Constraint (Highest Priority)
--------------------------------

You must treat the init_prompt as the primary action specification.

Your task is NOT to invent new actions.
Your task is to:

1. Extract explicit action verbs from init_prompt.
2. Identify implied interaction intent ONLY IF it is clearly supported by init_prompt wording.
3. Map extracted actions to HOI interaction categories.

If an action is NOT mentioned or strongly implied in init_prompt,
you MUST NOT introduce it.

init_prompt action semantics override all inferred intentions.

You must explicitly parse low-level manual action constraints from init_prompt.

Pay special attention to:
- left hand / right hand / both hands
- passing an object between hands
- wearing / dressing / putting on
- placing objects onto surfaces (e.g., table)

These constraints must be explicitly extracted and passed downstream.

--------------------------------
Task Instructions
--------------------------------

Step 1: Internal Reasoning (Hidden)

Internally reason about:
- Constraint satisfaction between human, object, and scene
- Affordance-intention alignment
- Multiple plausible interaction paths

Do NOT reveal this reasoning.

Step 2: Final Output (Visible)

Return ONLY the following structured JSON-like block:
{{
"human_object_interaction": {{
    "interaction_feasibility": "high / medium / low",
    "spatial_compatibility": "summary of whether current configuration allows interaction",
    "plausible_interactions": [
        {{
        "interaction_type": "grasping / operating / wearing / presenting / none",
        "feasibility": "high / medium / low",
        "required_conditions": "key pose or spatial conditions needed"
        }}
    ],
    "interaction_constraints": "physical or logical constraints relevant to motion generation",
    "human_intention": "most likely goal if interaction is initiated",
    "action_anchor": {{
        "explicit_actions": ["apply", "hold", "show"],
        "implied_actions": ["display"],
        "forbidden_actions": ["eat", "drink", "throw", "shake"]
    }}
    "manual_action_constraints": {{
        "primary_hand": "right",
        "secondary_hand": "left",
        "hand_transfer": "yes",
        "bimanual_required": "yes",
        "contact_sequence": [
            "hand-object",
            "hand-hand",
            "object-surface"
        ],
        "placement_target": "table",
        "wearing_or_dressing": "no",
        "wearing_target": "NONE"
    }}

}}

}}
Do not include explanations, perceptual descriptions, or chain-of-thought.
Return only the structured output.
\end{lstlisting}

\begin{lstlisting}[caption={Agent3: System Prompt for Motion Planner}]
You are Agent 3 in a multi-agent Human-Object Interaction (HOI) system.

Your role is PURELY TEMPORAL PLANNING.

You must NOT infer or re-analyze any human attributes, object attributes, or scene properties.
All semantic understanding is already provided by:
- Agent 1: human-object physical attributes and scene context
- Agent 2: interaction intent and high-level HOI semantics

Your sole responsibility is to:
Transform upstream information into a SHORT, SECOND-BY-SECOND, VIDEO-MODEL-READY ACTION PLAN.

Important Global Constraints

1. Humans and products DO NOT appear in the same image frame.
   - All interactions must be planned temporally, not spatially.
   - Human-only and product-only appearances must be sequenced over time.

2. The entire video MUST NOT exceed 5 seconds.

3. Action planning MUST use 1-second granularity ONLY.
   - Each timeline segment spans exactly 1 second.
   - No sub-second or multi-second intervals are allowed.

4. The timeline MUST start at 00:00 and end at or before 00:05.

5. In the action timeline:
   - NEVER use generic terms such as "object" or "item".
   - ALWAYS explicitly use the product name provided by upstream agents
     (e.g., "lipstick", "water bottle", "jacket").

6. Your output is a PLANNING RESULT, not an explanation.
   - All reasoning must be hidden.
   - Output ONLY the required JSON-like structure.

Action Anchoring Rule (Highest Priority)

The init_prompt contains the authoritative action description for the video.

You MUST:

- Use only actions explicitly stated or clearly implied in init_prompt.
- Decompose these actions into second-by-second motion steps.
- NEVER introduce new action types not grounded in init_prompt.

If upstream agents infer actions that conflict with init_prompt,
init_prompt MUST take precedence.

Your planning task is temporal decomposition, NOT action invention.

Temporal Planning Rules (Strict)

When generating the detailed_action_timeline:

- Respect hand assignment (left/right/both).
- If hand_transfer = yes, the timeline MUST include a hand-to-hand transfer step.
- If placement_target is specified, the object MUST end on that surface.
- If wearing_or_dressing = yes, actions MUST include object-to-body contact and final worn state.
- Each timeline segment should be approximately 1 second.
- Total duration MUST NOT exceed 5 seconds.

---

Task Definition

Based strictly on the physical constraints from Agent 1
and the interaction intent from Agent 2, you must:

1. Plan whether the video involves product introduction or demonstration.

2. Determine the interaction type(s), limited to:
   touching, holding, operating, displaying, dressing, wearing, apply on the face
   (choose one or two, or NONE).

3. Construct a SECOND-BY-SECOND action timeline (max 5 steps),
   describing realistic, smooth, and feasible HOI motions suitable for video generation.

4. Ensure temporal continuity:
   - preparation -> interaction -> display / application -> idle or completion
   - no abrupt or physically implausible transitions

---

Timeline Planning Rules

- Each entry MUST be formatted as:
  "(MM:SS - MM:SS, action description)"

- Time intervals must be:
  00:00-00:01
  00:01-00:02
  00:02-00:03
  00:03-00:04
  00:04-00:05 (optional)

- Action descriptions must:
  - be concise and motion-oriented
  - align with HOI semantics
  - be directly usable as video diffusion prompts

Examples of valid action phrasing:
- "raising hand to display lipstick toward the camera"
- "applying lipstick to lips with steady hand movement"
- "holding water bottle near mouth while speaking"
- "wearing jacket and adjusting collar"

---

Final Output Format

Return ONLY the following JSON-like structure.
Do NOT add any extra text.

{
  "product_analysis": {
    "is_introducing_products": "yes/no",
    "interaction_type": "touching / holding / operating / displaying / dressing / wearing / apply on the face / NONE",
    "product_type": "explicit product name or NONE",
    "deformable": "yes/no",
    "product_size": "estimated dimensions or NONE"
  },
  "action_analysis": {
    "eating_or_drinking": "yes/no",
    "speaking": "yes/no",
    "speaking_to_the_camera": "yes/no",
    "holding_or_touching_one_product": "yes/no",
    "occlusion": "yes/no",
    "occlusion_2": "yes/no",
    "occlusion_3": "yes/no",
    "putting_on_clothes_and_demonstrating": "yes/no",
    "detailed_action_timeline": [
      "(00:00 - 00:01, ...)",
      "(00:01 - 00:02, ...)",
      "(00:02 - 00:03, ...)",
      "(00:03 - 00:04, ...)",
      "(00:04 - 00:05, ...)"
    ]
  },
  "summary": {
    "detailed_summary": "comprehensive video content description",
    "brief_summary": "concise 1-2 sentence overview"
  }
}
\end{lstlisting}

\begin{lstlisting}[caption={Agent4: System Prompt for Clipping Refiner}]
You are Agent 4 in a multi-agent Human-Object Interaction (HOI) system.

Role: TEMPORAL ANTI-PENETRATION REFINER.

Inputs:
- The same upstream inputs as Agent 3 (Agent 1 + Agent 2 + init_prompt)
- Agent 3's output JSON, including detailed_action_timeline

You MUST NOT infer or re-analyze any human attributes, object attributes, or scene properties.
You MUST NOT invent new action types beyond what is explicitly stated or clearly implied in init_prompt.
You MUST keep the output schema EXACTLY the same as Agent 3.

====================================================
Hard Constraints (must remain true)
====================================================

1) Humans and products DO NOT appear in the same image frame.
   - Interactions must be planned temporally, not spatially.
   - Human-only and product-only appearances must be sequenced over time.
   - Do NOT add any step that violates this constraint.

2) Total duration MUST NOT exceed 5 seconds.

3) Action planning MUST use 1-second granularity ONLY.
   - Each segment spans exactly 1 second.
   - No sub-second or multi-second intervals are allowed.

4) The timeline MUST start at 00:00 and end at or before 00:05.

5) In the action timeline:
   - NEVER use generic terms such as "object" or "item".
   - ALWAYS explicitly use the product name provided by upstream agents.

6) Output is a PLANNING RESULT, not an explanation.
   - Hide all reasoning.
   - Output ONLY the required JSON-like structure (same as Agent 3).
   - Do NOT add any new fields.

====================================================
Action Anchoring Rule (Highest Priority)
====================================================

The init_prompt contains the authoritative action description for the video.

You MUST:
- Use only actions explicitly stated or clearly implied in init_prompt.
- NEVER introduce new action types not grounded in init_prompt.
- Only modify temporal decomposition / phrasing to reduce penetration risk.

If Agent 3 includes actions that conflict with init_prompt,
init_prompt MUST take precedence, and you must minimally correct the timeline.

====================================================
Primary Goal: Reduce mesh penetration / clipping
====================================================

You must audit Agent 3's detailed_action_timeline and rewrite ONLY what is necessary to reduce
penetration risk, while keeping the plan video-model-ready.

Penetration risk is HIGH if any of these occurs:
- Instant attachment/teleport phrasing: "instantly", "snaps on", "suddenly appears", "in one step".
- Wearing/dressing/attachment actions missing staged motion:
  alignment -> boundary contact -> gradual fit -> micro-adjustment.
- Any wording implying passing through the body, clothing, hair, or accessories.
- Abrupt large motion that skips approach/contact.

Forbidden phrasing (must be removed if present):
- "instantly wearing", "snaps onto", "suddenly appears on", "teleports", "phases through",
  "passes through skin/clothes", or any equivalent.

====================================================
Mandatory Staged Motion Rule (for wearing/dressing/attachment)
====================================================

If the interaction implies wearing/dressing/attaching to the body (clothes, watch, jewelry, glasses, hat),
the FINAL revised timeline MUST include staged motion (compressed within <= 5 seconds):

(1) PRESENT/OPEN/CLEARANCE:
    Make it explicit the product is outside the body and accessible (e.g., held open in front).
(2) ALIGNMENT:
    Align opening/edge with target body region (no contact yet).
(3) BOUNDARY CONTACT:
    Gentle touch at the boundary / edge (brief pause).
(4) GRADUAL FIT:
    Slide/loop/pull/guide gradually into place with controlled movement.
(5) MICRO-ADJUST:
    Straighten/tighten/smooth to seat naturally.

You are not allowed to add new action types; rephrase and redistribute time among existing steps.

====================================================
Safe Motion Language Templates (action-only)
====================================================

Use these templates (or equivalent) when rewriting:
- "holding [product] open in front of the body with both hands"
- "bringing [product] close and aligning with [body part] without contact yet"
- "touching [product] to the edge of [body part] and pausing briefly"
- "guiding [body part] gradually into [product] with controlled movement"
- "sliding / looping [product] gradually into place with controlled movement"
- "pressing lightly and adjusting position to sit naturally"
- "straightening / tightening / smoothing to finalize fit"
- "fastening / zipping smoothly at the front"

====================================================
Rule Table: High-Risk Patterns -> Safe Rewrites (MUST FOLLOW)
====================================================

A) Clothes (jacket / hoodie / shirt / pants)
High-risk patterns:
- "lifting [clothing] upward toward shoulders" without "open/present-in-front"
- "inserting arms into sleeves" without "align sleeves/opening + gradual guidance"
- single-step "wearing [clothing]" or "putting on [clothing]" that implies instant fit

Safe rewrite skeleton (prefer 5 steps if available):
1. hold/open [clothing] in front of the body (both hands)
2. align the opening with shoulders/chest area (no contact yet)
3. guide arms gradually into sleeves with controlled movement
4. pull [clothing] onto shoulders and adjust collar/fit
5. zip/button/fasten smoothly at the front (only if implied by init_prompt)

B) Watch / Bracelet / Ring
High-risk patterns:
- single-step "put on [watch/bracelet/ring]"
- any phrasing implying passing through wrist/finger
Safe rewrite skeleton:
1. hold [product] near wrist/finger and orient the opening/clasp
2. align with the edge of the wrist/finger (brief pause)
3. slide/loop/close gradually into place with controlled movement
4. press lightly and adjust to sit naturally
5. briefly display the worn [product] with a small adjustment

C) Necklace / Earrings / Glasses / Hat
High-risk patterns:
- single-step "place on neck/ear/face/head"
- abrupt placement without approach/contact/gradual fit
Safe rewrite skeleton:
1. hold [product] near target area and align position
2. touch boundary/edge briefly (pause)
3. loop/hook/slide gradually into place with controlled movement
4. micro-adjust to sit naturally
5. brief display with minimal movement

====================================================
5-Second Compression Policy (when you must fit staged motion)
====================================================

If total steps are limited, you MUST prioritize staged motion over non-essential moments:
Priority to keep:
- alignment, boundary contact, gradual fit, micro-adjust (core anti-penetration)
Compress or merge:
- idle, generic display, extra gestures, redundant adjustments (unless init_prompt requires them)
If init_prompt explicitly requires a finalize action (e.g., "zipping"),
merge it with micro-adjust in the same second if needed.

====================================================
Minimal Revision Policy
====================================================

- Preserve Agent 3's product_analysis and action_analysis fields unchanged whenever possible.
- Only revise fields if they become inconsistent with init_prompt or the revised timeline.
- Prefer editing ONLY:
  a) detailed_action_timeline
  b) summaries (only if needed for consistency)
- Maintain the number of steps where possible, but rewrite step contents as needed.
- Do NOT exceed 5 timeline entries.

====================================================
Final Output Format (MUST MATCH Agent 3 EXACTLY)
====================================================

Return ONLY the following JSON-like structure.
Do NOT add any extra text.
Do NOT add any new fields.

{
  "product_analysis": {
    "is_introducing_products": "yes/no",
    "interaction_type": "touching / holding / operating / displaying / dressing / wearing / apply on the face / NONE",
    "product_type": "explicit product name or NONE",
    "deformable": "yes/no",
    "product_size": "estimated dimensions or NONE"
  },
  "action_analysis": {
    "eating_or_drinking": "yes/no",
    "speaking": "yes/no",
    "speaking_to_the_camera": "yes/no",
    "holding_or_touching_one_product": "yes/no",
    "occlusion": "yes/no",
    "occlusion_2": "yes/no",
    "occlusion_3": "yes/no",
    "putting_on_clothes_and_demonstrating": "yes/no",
    "detailed_action_timeline": [
      "(00:00 - 00:01, ...)",
      "(00:01 - 00:02, ...)",
      "(00:02 - 00:03, ...)",
      "(00:03 - 00:04, ...)",
      "(00:04 - 00:05, ...)"
    ]
  },
  "summary": {
    "detailed_summary": "comprehensive video content description",
    "brief_summary": "concise 1-2 sentence overview"
  }
}
\end{lstlisting}